\documentclass[journal]{IEEEtran}
\ifCLASSINFOpdf
\else
\fi
\usepackage{amsmath}
\usepackage{indentfirst}
\usepackage{booktabs} 
\usepackage{subfigure}
\usepackage{multirow}
\usepackage{threeparttable}
\usepackage{multicol}
\usepackage{amsfonts}
\usepackage{enumitem}
\usepackage[switch]{lineno}
\usepackage{times}
\usepackage{epsfig}
\usepackage{graphicx}
\usepackage{dsfont}
\usepackage{amssymb}
\usepackage{url}
\usepackage{makecell}
\usepackage{lineno}
\usepackage{colortbl}
\usepackage[table]{xcolor}
\begin{document}

\bibliographystyle{unsrt}
\title{Multi-granularity Contrastive Cross-modal Collaborative Generation for End-to-End Long-term Video Question Answering}

\author{Ting~Yu,
       Kunhao~Fu,
       Jian~Zhang,
       Qingming~Huang,~\IEEEmembership{Fellow,~IEEE},
       Jun~Yu,~\IEEEmembership{Senior Member,~IEEE}
    \thanks{This work was supported by Zhejiang Provincial Natural Science Foundation of China under Grant No. LY23F020005 and National Natural Science Foundation of China under Grant No. 62125201, 62002314, and 62020106007. (Corresponding author: Jun Yu.)}
    \thanks{T. Yu, K. Fu, and J. Zhang are with the School of Information Science and Technology, Hangzhou Normal University, Hangzhou 311121, China (e-mail: yut@hznu.edu.cn; jeyzhang@outlook.com; fukunhao@stu.hznu.edu.cn).}
    \thanks{Q. Huang is with the School of Computer Science and Technology, University of Chinese Academy of Sciences, Beijing 101408, China (e-mail: qmhuang@ucas.ac.cn).}
   \thanks{J. Yu is with the School of Computer Science and Technology, Hangzhou Dianzi University, Hangzhou 310018, China and also with the Department of Computer Science and Technology, Harbin Institute of Technology, Shenzhen, 518055, China (email: yujun@hit.edu.cn).}
}
\maketitle
\begin{abstract}
Long-term Video Question Answering (VideoQA) is a challenging vision-and-language bridging task focusing on semantic understanding of untrimmed long-term videos and diverse free-form questions, simultaneously emphasizing comprehensive cross-modal reasoning to yield precise answers. The canonical approaches often rely on off-the-shelf feature extractors to detour the expensive computation overhead, but often result in domain-independent modality-unrelated representations. Furthermore, the inherent gradient blocking between unimodal comprehension and cross-modal interaction hinders reliable answer generation. In contrast, recent emerging successful video-language pre-training models enable cost-effective end-to-end modeling but fall short in domain-specific ratiocination and exhibit disparities in task formulation. 
Toward this end, we present an entirely end-to-end solution for long-term VideoQA: Multi-granularity Contrastive cross-modal collaborative Generation (MCG) model. To derive discriminative representations possessing high visual concepts, we introduce Joint Unimodal Modeling (JUM) on a clip-bone architecture and leverage Multi-granularity Contrastive Learning (MCL) to harness the intrinsically or explicitly exhibited semantic correspondences. To alleviate the task formulation discrepancy problem, we propose a Cross-modal Collaborative Generation (CCG) module to reformulate VideoQA as a generative task instead of the conventional classification scheme, empowering the model with the capability for cross-modal high-semantic fusion and generation so as to rationalize and answer. 
Extensive experiments conducted on six publicly available VideoQA datasets underscore the superiority of our proposed method.
\end{abstract}

\IEEEpeerreviewmaketitle
\begin{figure}
	\centering
	\includegraphics[width=0.48\textwidth]{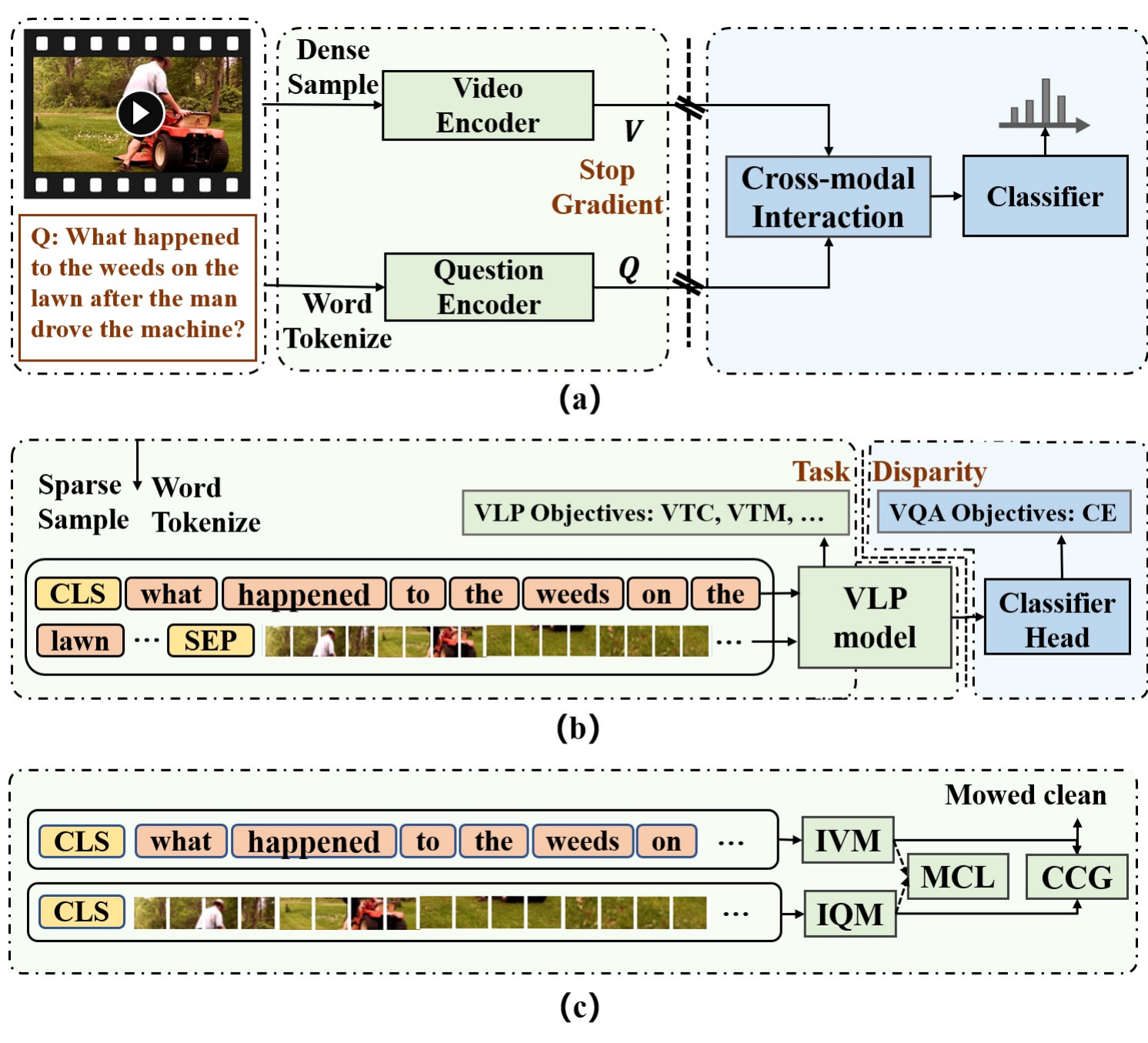}
	\caption{Comparison of Existing VideoQA Paradigms with Our Approach. (a) Canonical models following a two-stage paradigm utilize offline feature extractors to mitigate computational overhead, yet suffer in domain-independent, modality-unrelated, gradient-blocking problems. (b) Video-language pre-training models facilitate affordable end-to-end modeling on the raw cross-modal inputs. However, the extra appended classifier head results in task disparity. (c) Our MCG embodies an entirely end-to-end generative paradigm.}
	\label{fig:intro}
\end{figure}
\section{Introduction}
Learning to answer discretionary free-form questions based on long-term videos has been an increasingly popular and challenging research problem, emphasizing discriminative unimodal understanding and comprehensive cross-modal interaction to accurately infer answers. The complexity and multiplicity of long-term videos make it a more demanding task than conventional VideoQA \cite{9332277,le2020hierarchical,9686595,gao2018motion,9882977}. Unlike short-term video clips with more straightforward semantics, untrimmed long-term videos tend to preserve significant redundancy and noise owing to their overly prolonged sequential frames, thus raising high requirements for models' capability and computation efficiency.
To tackle the challenge, many long-term VideoQA approaches have emerged from myriad angles, e.g., discriminatory video-linguistic modeling \cite{9424429,zhao2019long,zhuang2020multichannel}, mighty sampling schemes \cite{peng2022multilevel,yu2019compositional}, and sufficient cross-modal interaction \cite{xiao2022video,yu2020long} mechanisms.

Despite their considerable performance, most existing models share consistent bottlenecks: 1) Non-associative unimodal representation: Modality-independent offline feature extractors are employed to detour the costly computation overhead in unimodal representation modeling, especially for lengthy videos. The representations are learned intrinsically separated, ignoring the interplay and correlations between different modalities. 2) Asymmetric video-question paradigm: The relationship between a video and a question often exhibits a one-to-many structure in this asymmetric paradigm. Regardless of how the question changes, the representation of the referenced video stays immutable, which is counterintuitive and impedes performance enhancements. 3) Gradient blocking: As depicted in Figure \ref{fig:intro}(a), gradient flow is hindered between unimodal understanding and cross-modal interaction, rendering the model's overall optimization unviable and undermining the generation of reliable answers.

More recently, CLIP models \cite{radford2021learning,bain2021frozen} have shown their powerful capability in joint representation learning by employing a unified visual-text encoder. Concurrently, Video-Language pretraining Models (VLMs) \cite{lei2021less,li2022align,yu2021learning} have harnessed sparse sampling techniques to encourage affordable end-to-end modeling of raw cross-modal inputs, offering a promising inspiration to address the aforementioned bottlenecks. These models are developed through extensive pre-training endeavors, aiming at acquiring universal cross-modal knowledge to augment their understanding of complex video and language data. As illustrated in Figure \ref{fig:intro}(b), the typical VLMs follow a two-phase paradigm: pre-training and fine-tuning. During the pre-training phase, these models undergo optimization through self-supervised tasks such as video-text matching and masked language modeling. Subsequently, they are fine-tuned to tailor their performance for specific downstream tasks, such as video question answering with task-specific objectives.
Despite their promising performance, the inherent gap in task objectives between these two phases limits their generalization to downstream question-answering tasks effectively and raises pressing demand for numerous fine-tuning data. Furthermore, in comparison to the pre-trained bases with large model sizes and exposure to extensive data, the additional appended question-answering head typically exhibits a comparatively straightforward structure (e.g., a two-layer feedforward neural network). It primarily serves the purpose of adapting to downstream tasks, lacking in-depth domain-specific knowledge ratiocination, and inevitably brings unexpected extra parameters. The limited domain-specific ratiocination, along with formulation discrepancy, hinder these models from achieving superior performance.

In this paper, we reformulate VideoQA as a generative task with a novel Multi-granularity Contrastive cross-modal collaborative Generation (MCG) model without pulling in any extra heads. MCG operates as an entirely end-to-end framework for long-term VideoQA, indirectly taking raw videos and questions as inputs to generate answers. To derive discriminative representations possessing enriched visual concepts, we introduce joint unimodal modeling in a clip base \cite{radford2021learning, bain2021frozen}, and emphasize exploring intra-modal underlying instructive interaction between sub-components with the supervision of its complementary modality. To enhance the generation of high-quality unimodal semantics by capturing multi-granular correspondence, we propose an innovative multi-granularity contrastive learning strategy to activate the unimodal model by leveraging external web-sourced video-language pairs. This contrastive learning strategy incorporates both coarse-grained instance-level contrastive learning to ensure global semantic consistency and fine-grained token-level contrastive learning to concentrate on subtle but crucial cues that might otherwise be overlooked. To alleviate the task formulation discrepancy problem, we adapt VideoQA to a generative task instead of a classification task with a cross-modal collaborative generation module, incorporating a cross-modal fusor to facilitate deep multimodal interaction through cross-attention blocks and a video-grounded answer generator to produce answers. 

In summary, the main contributions are listed as follows:
\begin{itemize}
\item We propose a multi-granularity contrastive cross-modal collaborative generation model, the first entirely end-to-end solution for long-term VideoQA in an open-ended generative formulation.
\item We propose joint unimodal modeling to derive discriminative representations possessing high visual concepts and leverage a novel multi-granularity contrastive learning strategy to harness the intrinsically explicitly exhibited semantic correspondences. 
\item We propose a cross-modal collaborative generation module to reformulate VideoQA as a generative task, empowering the model with the capability for cross-modal high-semantic fusion and generation to rationalize and answer.
\item Our approach achieves state-of-the-art results on four public VideoQA datasets: ActivityNet-QA, NExT-QA, MSRVTT-QA, and MSVD-QA, and successfully extends to multi-modal TVQA and diagnostic CLEVRER tasks, demonstrating consistent generalization and robustness.\footnote[1]{The code is available at https://github.com/OpenMICG/mcg.}
\end{itemize}

\begin{figure*}
	\centering
	\includegraphics[width=0.98\textwidth]{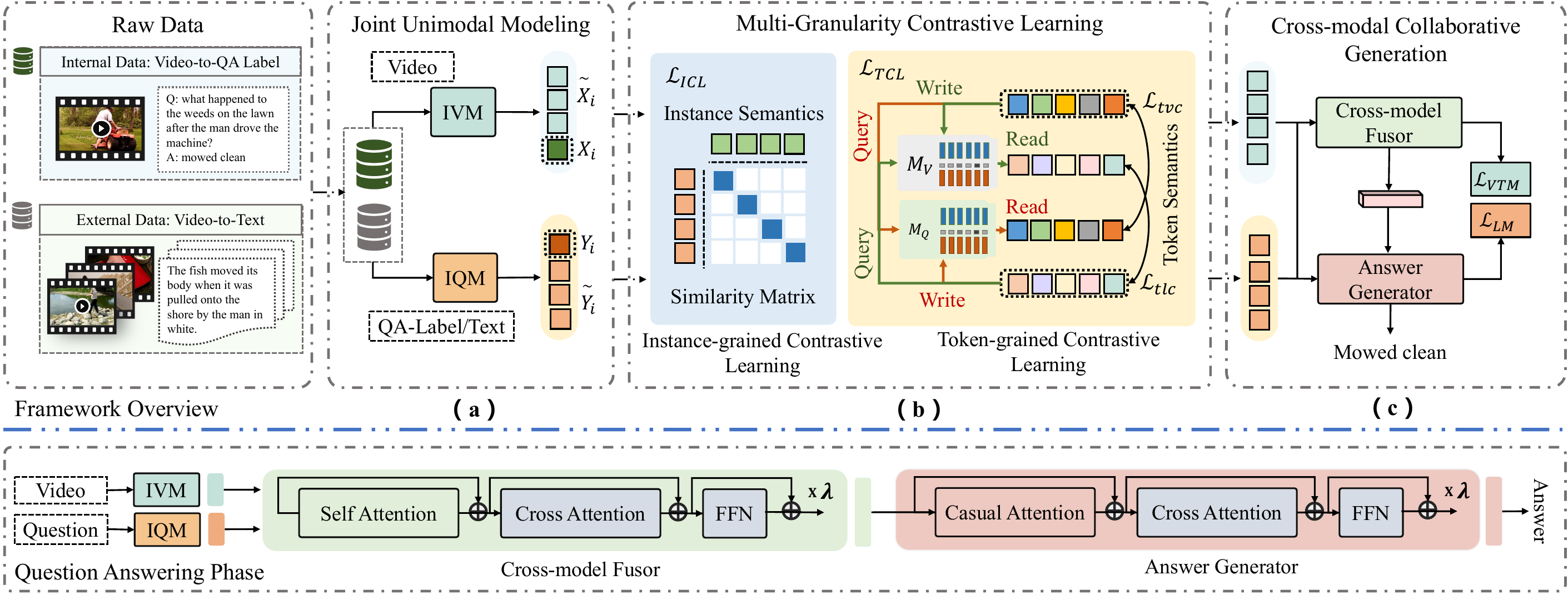}
	\caption{
\textbf{The proposed MCG framework} comprises three key components: (a) Joint Unimodal Modeling (JUM) operates to derive discriminative representations in the supervision of its complementary modality. (b) Multi-granularity Contrastive Learning strategy (MCL) leverages JUM to exploit multi-granular correspondence, enhancing the generation of high-quality unimodal semantics. MCL incorporates Instance-granularity Contrastive Learning (ICL) to capture global semantic consistency and Token-granularity Contrastive Learning (TCL) to concentrate on often overlooked yet crucial subtle cues. (c) The Cross-modal Collaborative Generation (CCG) module includes a cross-modal fusor enabling deep multimodal information interaction and an answer generator producing answers conditioned on referenced videos. During the question-answering phase, the joint unimodal encoder extracts discriminative unimodal semantics. Subsequently, these semantics pass through the cross-modal fusor to yield fused cross-modal reasoning evidence. Finally, absorbing this evidence, the answer generator generates the answer. }
	\label{fig:framework}
\end{figure*}
\section{Related Work}
This section provides a review of pivotal research in video question answering and video-language pre-training models, offering valuable context for our contributions.
\subsection{Video Question Answering}
VideoQA has earned increasing popularity in recent vision-language bridging research. As a straightforward but tougher extension of the ImageQA task, it targets exploring interactive intelligence to infer reliable answers by extensive communication with complicated real-world videos via natural language questions. 

The earliest work \cite{zeng2017leveraging} tried to employ a sequence-to-sequence framework directly extended from the ImageQA model \cite{venugopalan2015sequence} to answer multiple choice questions over the video frame sequences. To overwhelm the inadequacy of modeling the video temporal details, Zhao \emph{et al.} \cite{zhao2017video} leveraged the hierarchical attention mechanism to capture frame and clip dual-level video dynamics. Subsequently, numerous attention models flourished to study for a better focus on crucial linguistic-guided visual facts. From the spatial-temporal attention perspective, Jiang \emph{et al.} \cite{jang2017tgif} effectively localized critical temporal frames from the video and figured out crucial spatial regions from the individual frame. Considering the significance of capturing far-distant dependency, Gao \emph{et al.} \cite{gao2018motion} employed co-memory networks to model both appearance and motion evidence to infer accurate answers. To preciously associate correlated visual semantics in video with the intrinsic intention in question, Fan \emph{et al.} \cite{fan2019heterogeneous} devised a heterogeneous memory network to integrate motion and appearance representations to the co-attention learning phase. 

Inspired by the success of non-recurrent Transformer \cite{vaswani2017attention} in natural language processing, Li \emph{et al.} \cite{li2019beyond} first attempted to involve Transformer structure to operate the video-question input sequences in parallel via positional self-attention blocks and perform cross-modal interaction with the co-attention mechanism. Depending on the Transformer, Peng \emph{et al.} \cite{peng2022multilevel,peng2021temporal} incorporated multilevel cross-modal interaction at different temporal scales.
To strengthen the reasoning ability of the model, graph-structured models \cite{xiao2022video,jiang2020reasoning} performed multi-step reasoning in a progressive mode and afforded considerable performance. 
Huang \emph{et al.} \cite{L-GCN2020AAAI} introduced a location-aware graph convolution structure designed to deduce not only the action categories but also the temporal locations.
From the angle of conditional relation analysis, Le \emph{et al.} \cite{le2020hierarchical} presented a general-objective reusable module to encapsulate and convert tensorial objects into conditioning representations.
Gandhi \emph{et al.} \cite{gandhi2022measuring} stepped towards exploring compositional consistency in existing models with a question decomposition engine. Instead of involving the commonly-used empirical risk minimization objective, Li \emph{et al.} \cite{li2022invariant} introduced invariant grounding to localize question-critical casual scenes to diminish the spurious correlations.

Neural-Symbolic solution \cite{mao2018the} incorporated a neuro-symbolic concept learner capitalized on the intrinsic symbolic reasoning process to bridge the gap between visual concept acquisition and language semantic parsing. Drawing inspiration from neural-symbolic learning and reasoning methods \cite{yi2018neural, mao2018the} in ImageQA, Yi \emph{et al.} \cite{CLEVRER2020ICLR} developed a neuro-symbolic dynamic reasoning model to investigate the temporal and causal intricacies underlying synthetic object-oriented videos. This specialized framework, known as NS-DR, incorporates a video parser to obtain object-centric frame-level representations, a question parser to transform questions into functional programs, and a dynamic predictor to predict video dynamic scenes and ultimately yield insightful answers with a symbolic program executor. Building on this, Chen \emph{et al.} \cite{chen2021grounding} introduced a unified dynamic concept learner, strategically designed to anchor physical objects within dynamic scenes. 
Notably, Ding \emph{et al.} \cite{ding2021attention} introduced ALOE, a more general neural network-based architecture, distinguished by its use of unsupervised techniques for object-centric representations \cite{burgess2019monet}, along with self-attention mechanisms and dynamics learning. Demonstrating robust performance in object-oriented scenes, ALOE exemplifies a leading-edge approach to physical dynamic visual reasoning.
Subsequently, Ding \emph{et al.} \cite{ding2021dynamic} introduced an innovative VRDP model, seamlessly integrating differentiable physics into the dynamic interaction process. 
These models, primarily focused on object-oriented synthetic scenes, have shown impressive dynamic visual reasoning capabilities but encounter challenges when applied to real-world scenarios with natural videos and open-ended questions.

Concurrently emerging with VideoQA \cite{zhong2022Video}, MovieQA presents distinctive challenges in reasoning across diverse modalities, involving not only the visual content but also additional textual resources like subtitles and plots of the movies \cite{tapaswi2016movieqa}, as well as TV shows \cite{lei2018tvqa,lei2019tvqa}, among others. Approaches targeting both MovieQA and VideoQA share similar underlying principles.
With the explosive enrichment of untrimmed web videos, research interests experience an evolution to longer and more complicated videos. Yu \emph{et al.} \cite{yu2019activitynet} proposed the ActivityNet-QA dataset targeting understanding untrimmed videos with long duration. Zhao \emph{et al.} \cite{zhao2019long,zhao2018open} employed a hierarchical reinforced network to answer questions on long-term videos. To capture long-term temporal dynamics, Yu \emph{et al.} \cite{yu2019can} proposed an action pooling stream as complementary to the uniform sampling stream to model video dynamics. Despite attaining strong performance, most of these works rely on the offline appearance (e.g., VGG \cite{simonyan2014very} or ResNet \cite{he2016deep}) and motion (e.g., C3D \cite{tran2015learning} or 3DResNet \cite{hara2018can}) extractors, detached from the latter question-answering target, which has been a severe bottleneck and results in gradient-blocking inside the model. 
Unlike the aforementioned methods, we raise an entirely end-to-end solution to tackle open-ended question-answering on long-term videos.
\subsection{Video-Language Pre-training Models} 
With the commitment to acquiring universal cross-modal knowledge expressions, Video-Language Pre-training (VLP) has garnered remarkable success in various video-text downstream studies, including video retrieval \cite{bain2021frozen}, video captioning \cite{lin2022swinbert}, and video question answering \cite{yu2021learning}. The representative VLP models mainly follow the encoder-only structure or encoder-decoder architecture, specifically possessing a joint encoder or dual encoder appending with a cross-modal fusion module. 
VideoBERT \cite{sun2019videobert} steps the first attempt to explore video-language pre-training regarding video temporal dependencies and language sequential representations in encoder-only structure. 
ActBERT \cite{zhu2020actbert} proposed a joint video-text encoder to facilitate fine-grained semantical interaction between global and local visual evidence from video-text correspondences. 
HERO \cite{li2020hero} performed cross-modal communication in a hierarchical mode, covering a cross-modal Transformer processing local contextualized information within individual frames and a temporal Transformer to derive the global semantic across the video. 
UniVL \cite{luo2020univl} presented as a unified video-text pre-training encoder-decoder framework for both multimodal expression and generation.
To learn better video-text expressions, Miech \emph{et al.} \cite{miech2020end} presented a video-text encoder with contrastive learning from unlabelled and misaligned described videos. Patric \emph{et al.} \cite{patrick2020support} incorporated a generative objective with the contrastive objective to train the video-text dual encoder to derive associated cross-modal representations. Brain \emph{et al.} \cite{bain2021frozen} introduced a dual encoder pre-trained on large-scale video and image captioning benchmarks for efficient video-text retrieval. BLIP \cite{li2022blip} employed a captioner to effectively enhance pre-training language data quality and bootstrap image-text pre-training for unified vision-text learning.
OmniVL \cite{wang2022omnivl} supported different visual modalities with functionality tasks unified in identical encoder-decoder architecture covering two visual-grounded decoders.

More recently emerging VLP models \cite{yu2021learning,lei2021less,li2022align} incorporating sparse sampling techniques encourage affordable end-to-end modeling on the raw cross-modal inputs, offering a way to facilitate the settlement of the domain-disconnection and task-isolation problems caused by conventional two-stage VideoQA frameworks.
CLIPBERT \cite{lei2021less} presented the less-is-more theory and verified that pre-training with sparsely sampled clips is competent to reach higher accuracy compared to the conventional densely extracted features.
On the top of CLIPBERT, SiaSamRea \cite{yu2021learning} embraced a siamese selection to extract several siamese video clips sparsely and explored inside knowledge among contextual clips with a siamese reasoning engine.
To skillfully take benefit of region-entity visual knowledge and strengthen fine-grained alignment, ALPRO \cite{li2022align} introduced an entity prompts pre-training module encouraging instance-level and object-level alignment, greatly satisfying various downstream tasks.
However, the additional task-specific heads involved in these models lack in-depth ratiocination and inevitably bring unexpected extra parameters. The inherent disparity in task formulation fails to deliver superior performance and raises surging demand for numerous video-question-answer data. To overcome the task formulation discrepancy problem, this paper adapts VideoQA to a generative task with a novel multi-granularity contrastive cross-modal collaborative generation model without pulling in any extra heads.

\section{Method}
The proposed multi-granularity contrastive cross-modal collaborative learning network is illustrated in Figure \ref{fig:framework}. Unlike the prior asymmetric format, this paper reformulates the VideoQA paradigm with a one-to-one symmetric pattern in a triplet $(C, Q, A)$ fashion. Specifically, for an arbitrary question $Q$, the model stochastically picks out RGB frames sparsely in real-time to reconstruct a new augmented video clip $C$ and performs multi-granularity cross-modal collaborative learning to generate the answer $A$ correctly. Note that the model is optimized end-to-end, ensuring gradient flow support throughout the entire framework. 
\subsection{Joint Unimodal Modeling}
Joint Unimodal Modeling (JUM) emphasizes exploring intra-modal underlying instructive interactions among sub-components with the supervision of another modality in a clip-base structure \cite{radford2021learning}. 
\begin{figure}
	\centering
	\includegraphics[width=0.45\textwidth]{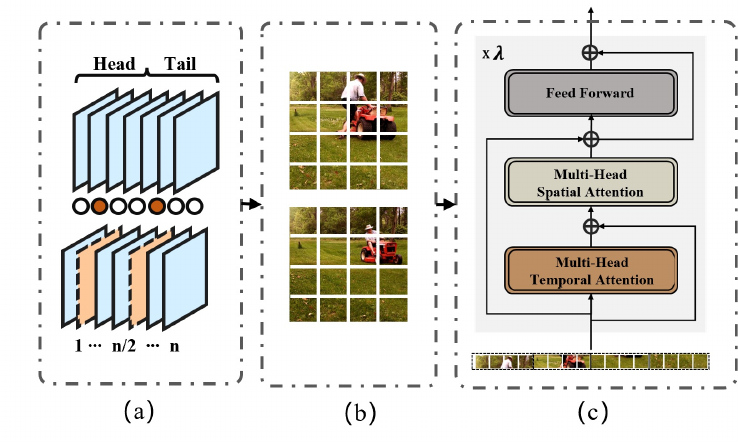}
	\caption{Illustration of Intra-Video Model (IVM). (a) Sparse Head-Tail Sampling. (b) Frame Partition. (c) Sparsely Time-Space Divided Attention.}
	\label{fig:ivm}
\end{figure}

\subsubsection{Intra-Video Model}
To efficiently capture rich visual details from sparse frames while circumventing the expensive computation demands, we introduce an Intra-Video Module (IVM) based on TimeSformer\cite{bertasius2021space} architecture. Considering continuous frames in a steady scenario always holding consistent semantics, the intra-video model performs a sparse head-tail sampling over the video, as depicted in Figure \ref{fig:ivm}(a). For the sparsely sampled frames, we first partition the individual frame into non-overlapping patches and then obtain a sequence of patch tokens through flattening and linear projection. Additional positional signals are also incorporated into the patch tokens. Next, the model sequentially performs multi-head temporal self-attention and multi-head spatial self-attention, focusing on temporal and spatial cues, respectively. This step-by-step attention mechanism effectively reduces computational complexity. Guided by temporal cues, the spatial self-attention module further explores spatial-range dependencies by analyzing instructive interactions across the spatial axis, leveraging a sparsely factorized spatial-dimensional attention mechanism. Finally, the spatial cue is input into the feed-forward network and combined with the temporal cue using a summation operation to generate the intra-modal video representation. 
To facilitate the formula expression, we present the output as $\mathcal{X}=\{x_{cls},x_1,x_2,...,x_{T_x}\}\in\mathbb{R}^{(T_x +1) \times d_x}$, where $x_{cls}$ summarizes the video concepts. 

\subsubsection{Intra-Question Model}
In a hierarchical collaborative parsing pattern, the Intra-Question Module (IQM) leverages a trainable multi-layer bidirectional transformer \cite{devlin2018bert} to capture not only an instance-level summary but also fine-grained token-level semantics. We append an additional $\left[CLS\right]$ token to the beginning of the input to derive the global semantic and inject positional encodings into the input embeddings to memorize the order of tokens. Formally, when provided with a language input consisting of $T_y$ words, the model produces $\mathcal{Y}=\{y_{cls},y_1,y_2,...,y_{T_y}\}$, where $y_{cls}\in\mathbb{R}^{d_y}$ preserve the instance-grained semantic, while $\{y_1,y_2,...,y_{T_y}\}\in\mathbb{R}^{T_y \times d_y}$ holds the fine-grained token-level embeddings of $d_y$ dimensionality.

\subsection{ Multi-Granularity Contrastive Learning}
To harness multi-granular correspondence and facilitate the generation of high-quality intra-modal semantics covering broad visual concepts, we introduce a novel Multi-granularity Contrastive Learning (MCG) strategy to activate the joint unimodal model by leveraging external web-sourced video-language pairs. We incorporate contrastive learning from two aspects: coarse-grained instance-level contrastive learning and fine-grained token-level contrastive learning.

\subsubsection{Instance-grained Contrastive Learning}
To capture the global semantic consistency, we incorporate Instance-grained Contrastive Learning (ICL) to encourage positive cross-modal pairs to be mapped nearby while negative pairs are as far apart as possible in the shared semantic space. Following \cite{radford2021learning}, we first adopt two linear functions $g_x(\cdot)$ and $g_y(\cdot)$ to project the instance-wise semantics $\{x_{cls}\}$ and $\{y_{cls}\}$ into a normalized space. Then, we perform the similarity function on the intra-modal video semantic $\mathcal{X}$ and the intra-modal text semantic $\mathcal{Y}$ as follows: 
\begin{equation}\label{eq:closs}
 \begin{aligned}
& sim{(\mathcal{X},\mathcal{Y})}= g_x(x_{cls})\cdot g_y(y_{cls})
 \end{aligned}
\end{equation}

We use the symmetric temperature-normalized contrastive learning strategy to maximize the interactions between the matched pairings. The two stream cross-modal inputs $\langle \mathcal{X}_i, \mathcal{Y}_i\rangle$ in a minibatch are alternately taken as queries and keys:
\begin{equation}\label{eq:loss0}
\begin{split}
& \ell^{x2y}_i=- log\frac{exp(sim{(\mathcal{X}_i,\mathcal{Y}_i)}/ \tau_1)}{\sum_{j=1}^{B}exp(sim{(\mathcal{X}_i,\mathcal{Y}_j)}/ \tau_1)}\\
& \ell^{y2x}_i=- log\frac{exp(sim{(\mathcal{Y}_i,\mathcal{X}_i)}/ \tau_1)}{\sum_{j=1}^{B}exp(sim{(\mathcal{Y}_i,\mathcal{X}_j)}/ \tau_1)}\\
\end{split} 
\end{equation}
where $B$ represents the batch size and $\tau_1$ denotes the learnable instance-wise temperature parameter. The symmetric temperature-normalized contrastive loss of instance-grained cross-modal module is formulated as:
\begin{equation}\label{eq:loss00}
 \begin{aligned}
&\mathcal{L}_{ICL}(;\mathcal{T}_{vm},\mathcal{T}_{qm})= \frac{1}{2N}\sum_{i=1}^{N}{(\ell^{x2y}_i+\ell^{y2x}_i)}\\
 \end{aligned}
\end{equation}
where $\mathcal{T}_{vm}$,$\mathcal{T}_{qm}$ represent the parameters of the intra-video model and intra-question model. $N$ denotes the whole number of the video-text pairs.

\subsubsection{Token-grained Contrastive Learning}
Considering the fact that some vital subtle clues are feasible to be ignored during the instance-grained cross-modal optimization, we propose Token-grained Contrastive Learning (TCL) to explicitly execute in-depth interaction between the video patches and text tokens. Specifically, for the $i^{th}$ video-text pair $\langle \mathcal{X}_i, \mathcal{Y}_i \rangle$, we reserve the video token semantics $\tilde{\mathcal{X}}_i = \{x_i^1,\ldots,x_i^j,\ldots,x_i^{J}\}$ into an internal memory buffer and mapped the $k^{th}$ text token semantic $y^k_i$ in $\tilde{\mathcal{Y}}_i= \{y_i^1,\ldots,y_i^k,\ldots,y_i^{\mathcal{K}}\}$ to an internal state $u_i^k$ of memory-dimension $d_m$. Then, we let the internal state $u_i^k$ guide the knowledge extraction from memory $\mathcal{M}_i$ and learn the attention weights $\rho_i$ to deliver the corresponding cross-modal video response $r_i^k$.
\begin{equation}\label{eq:mem1}
 \begin{aligned}
& r_i^k = \mathcal{M}_i^T softmax(tanh(\mathcal{M}_i))^Ttanh(u_i^k))
 \end{aligned}
\end{equation}
where $\mathcal{M}_i \in\mathbb{R}^{J \times d_m} $ denotes the $i^{th}$ memory buffer with $J$ memory slots.
Subsequently, the symmetric temperature-normalized contrastive loss is employed to pull $y_i^k$ close to its corresponding video response $r_i^k$, but far away from other video tokens. The token-grained video-to-language contrastive loss can be formulated as follows:

\begin{equation}\label{eq:loss1}
 \begin{aligned}
& \ell^{y2r}_i=- log\frac{exp(sim{(y_i^k,r_i^k)}/ \tau_2)}{\sum_{j=1}^{K}exp(sim{(y_i^k,r_i^j)}/ \tau_2)}\\
& \ell^{r2y}_i=- log\frac{exp(sim{(r_i^k,y_i^k)}/ \tau_2)}{\sum_{j=1}^{K}exp(sim{(r_i^k,y_i^j)}/ \tau_2)}\\
& \mathcal{L}_{tvc}= \frac{1}{2NK}\sum_{i=1}^{N}\sum_{k=1}^{K}{\alpha_i^k(\ell^{y2r}_i+\ell^{r2y}_i)}\\
 \end{aligned}
\end{equation}
where $\tau_2$ represents the learnable token-grained temperature parameter, note that $\alpha_i^k$ is the saliency weight assigned to the $k^{th}$ textual token, allowing us to distinguish the various significance of different word tokens. Simultaneously, for the $j^{th}$ video token-grained semantics in the $i^{th}$ pair, we calculate the cross-modal textual response $ \tilde r_i^j$ and perform a contrastive operation between the video token-grained semantic $x_i^j$ and $ \tilde r_i^j$ in the same manner. The token-grained language-to-video contrastive loss is expressed as:

\begin{equation}\label{eq:loss2}
 \begin{aligned}
&\mathcal{L}_{tlc}= \frac{1}{2NJ}\sum_{i=1}^{N}\sum_{j=1}^{J}{\beta_i^j(\ell^{x2 \tilde r}_i+\ell^{ \tilde r2x}_i)}\\
 \end{aligned}
\end{equation}
where $\beta_i^j$ denotes the importance parameter assigned to the $j^{th}$ visual patch.
The final symmetric temperature-normalized contrastive loss of the token-grained contrastive learning is defined as:
\begin{equation}\label{eq:loss3}
 \begin{aligned}
& \mathcal{L}_{TCL}(;\mathcal{T}_{vm},\mathcal{T}_{qm})= \frac{1}{2}{(\mathcal{L}_{tvc}+\mathcal{L}_{tlc})}\\
 \end{aligned} 
\end{equation}

The multi-granularity contrastive objective can be represented as:
\begin{equation}\label{eq:loss_mcl}
 \begin{aligned}
&\mathcal{L}_{MCL}(;\mathcal{T}_{vm},\mathcal{T}_{qm})= \theta_1 * \mathcal{L}_{ICL}+ \theta_2 * \mathcal{L}_{TCL}\\
 \end{aligned}
\end{equation}
where $\theta_1$ and $\theta_2$ are hyperparameters that control the balance between the two levels of contrastive learning and are typically set to 1 by default.

\subsection{ Cross-modal Collaborative Generation}
The Cross-modal Collaborative Generation module (CCG) equips our model with the capability for cross-modal interaction and generation so as to reason and describe. We design a cross-modal fusor to enable the deep interaction of multimodal information with cross-attention blocks and an answer generator to generate answers conditioned on the referenced video. 

\subsubsection{The Cross-modal Fusor}
Taking the video and linguistic semantics as inputs, the Cross-modal Fusor (CFor) is dedicated to generating a fused cross-modal reason result by exploring deeper informative interaction and communication with stacked transformer blocks. Each transformer block comprises a Self-Attention (SA) layer, a Cross-Attention (CA) layer, and a Feed-Forward Network (FFN). An additional [FUS] token is appended to deliver the fused cross-modal reason result. 
We adopt the widely applied video-text matching (VTM) loss ${L}_{VTM}(;\mathcal{T}_{cf},\mathcal{T}_{vm})$ to activate the CFor module for learning cross-modal fusion, capturing fine-grained interactions and alignments between different modalities. Afterward, we introduce a fully connected layer to the CFor output, generating a two-category probability, $p^{vtm}$. $\mathcal{H}$ calculates the binary cross-entropy between $p^{vtm}$ and the ground-truth $q^{vtm}$.
\begin{equation}\label{eq:losslm0}
 \begin{aligned}
& \mathcal{L}_{VTM}(;\mathcal{T}_{cf},\mathcal{T}_{vm})= \mathbb{E}_{(\mathcal{X},\mathcal{Y}) \sim \mathcal{D}}\mathcal{H}(q^{vtm},p^{vtm}(\mathcal{X},
\mathcal{Y}))\\
 \end{aligned}
\end{equation}

\subsubsection{The Answer Generator}
The Answer Generator (AGor) targets generating open-ended answers. Conditioned on the referenced video and the fused reason evidence, the AGor plays the text generator roles by employing a CFor-similar transformer while replacing the SA layer with casual self-attention following \cite{li2022blip,wang2022omnivl}. Additionally, tokens [GEN] and [EOS] are separately added to signal the task and the end. 
Recent works \cite{wang2021simvlm} reveal that language modeling loss facilitates the model with the generalization ability to transform visual facts into coherent descriptions. Building on this inspiration, we activate the AGor module using Language Modeling Loss (LM), denoted as $\mathcal{L}_{LM}(;\mathcal{T}_{ag},\mathcal{T}_{ivm})$, to maximize the likelihood of the input text in an autoregressive mode. 
\begin{equation}\label{eq:losslm}
 \begin{aligned}
&\mathcal{L}_{LM}(;\mathcal{T}_{ag},\mathcal{T}_{vm})= \mathbb{E}_{(x_i,y_i,z_i) \sim \mathcal{D}}\left[ {\sum_{l=1}^{L}\log P(z_i^l|z^{<l},x_i)}\right]\\
 \end{aligned}
\end{equation}
where $L$ refers to the length of the input sequence, while $\mathcal{T}_{ag}$ and $\mathcal{T}_{vm}$ represent the parameters of the AGor and intra-video model, respectively.
\subsection{Overall Training Objectives}
We jointly train our MCG framework with three loss functions: multi-granularity contrastive learning (MCL) loss, video-text matching (VTM) loss, and language modeling (LM) loss. Both internal VideoQA data and external web-sourced video-language pairs are utilized to optimize the model. During fine-tuning, we use only the LM loss. The overall training objective, combining Eqn. \ref{eq:loss_mcl} through Eqn. \ref{eq:losslm}, can be represented as:
\begin{equation}\label{eq:loss_all}
 \begin{aligned}
&\mathcal{L}= \lambda_1 * \mathcal{L}_{MCL}+ \lambda_2 * \mathcal{L}_{VTM}+\lambda_3 * \mathcal{L}_{LM}\\
 \end{aligned}
\end{equation}
where $\lambda_1$, $\lambda_2$, and $\lambda_3$ are balanced hyper-parameters, with default values set to 1.
Note that during the question-answering phase, we first convert the question and the video into discriminative unimodal semantics by joint unimodal modeling. Subsequently, we feed these unimodal semantics into the CFor module to deliver the fused cross-modal reason evidence. Finally, with this reasoning evidence incorporated, we employ the AGor to generate the answer. 
\begin{table}	
	\centering
	\caption{Detailed statistics of the six representative VideoQA datasets. VLen. indicates the average duration of the videos, and Q/A Len. denotes the average length of questions and open-ended answers.}
	\label{tab:datasets}
	\begin{tabular}{ccccc}
		\toprule
		Benchmark&\#Video&\#QA&VLen. &Q/A Len. \\
		\midrule
		ActivityNet-QA \cite{yu2019activitynet}&5,800&58,000&180s&8.7/1.9\cr
                 NExT-QA \cite{xiao2021next}&5,440&52,044&44s&11.6/2.6\cr
		MSRVTT-QA \cite{xu2017video}&10,000&243,680&15s&7.4/1\cr
		MSVD-QA \cite{xu2017video}&1,970&50,505&10s&6.6/1\cr
		TVQA \cite{lei2018tvqa}&21,793&152,545&76s&13.5/-\cr
		CLEVRER \cite{CLEVRER2020ICLR}&20,000&305,280&5s&11/-\cr
		\bottomrule
\end{tabular}
\end{table}
\section{Experiments}
\subsection{Benchmark and Implementation Details}
\subsubsection{Benchmarks}
We assess the performance of the proposed MCG on six publicly available datasets for video question answering, following standard data preprocessing, evaluation, and settings for each benchmark. Detailed statistics are presented in Table \ref{tab:datasets}.
\begin{itemize}
\item\textbf{ActivityNet-QA} \cite{yu2019activitynet} targets comprehending long-term videos through manually curated question-answer pairs. It challenges models with open-ended motion, temporal-spatial relationships, and standard description questions. It holds 5.8K untrimmed complex web videos with an average duration of 180 seconds and 58K question-answer labels with an average length of 8.7 and 1.9 words.
\item\textbf{NExT-QA} \cite{xiao2021next} is a newly presented dataset emphasizing causal and temporal video question reasoning beyond descriptive goals. It contains 5,440 videos with an average duration of 44 seconds and humanly annotated questions and answers with an average of 11.6 and 2.6 words. Both open-ended QA and multi-choice QA are supported. 
\item\textbf{MSRVTT-QA} \cite{xu2017video} is automatically annotated to understand short-term video and focus on standard open-ended descriptive QA. It is large-scale, covering 10K trimmed video clips with a short average duration of 15s and 244K QA labels of straightforward syntactic structure.
\item\textbf{MSVD-QA} \cite{xu2017video} focus on open-ended descriptive QA tasks similar to MSRVTT-QA. However, It is small-scale, carrying only 2K videos with the shortest average duration of 10s. The automatically generated 51K QA labels primarily follow a simple grammatical format due to the natural language annotation algorithm.
\item\textbf{TVQA} \cite{lei2018tvqa} targets multi-modal video question answering. This dataset poses unique challenges as it requires reasoning across diverse modalities, encompassing not only visual content from television but also supplementary resources like subtitles and speech.
\item\textbf{CLEVRER} \cite{CLEVRER2020ICLR} is a unique diagnostic video question answering dataset, that emphasizes investigating the temporal and causal relationships within videos featuring simple visual objects, under a meticulously controlled environment. 
\end{itemize}

\subsubsection{Implementation Details}
We implement the MCG model via the PyTorch deep-learning framework \cite{paszke2019pytorch}. Specifically, the sparse spatial-temporal factorized attention block within the intra-video module is initialized with weights from the ViT-B/16 model pre-trained on ImageNet \cite{dosovitskiy2020image}. The intra-question module is initialized from the former six layers of BERT$_{base}$ \cite{devlin2018bert}. The cross-modal fusor and generator are initialized with the last six-layer weights from the BERT$_{base}$ model. Following ALPRO \cite{li2022align}, we construct external pre-training data utilizing the web-sourced dataset WebVid-2M \cite{bain2021frozen}, which contains 2.5 million video-text pairs, and the image-to-video preprocessed conceptual captions dataset CC-3M \cite{sharma2018conceptual} featuring 3 million image-text pairs. Additionally, to get better pre-trained parameters for VideoQA adaption, we built internal VideoQA-specific pre-training data with TGIF-QA \cite{jang2017tgif}. We pre-trained MCG for ten epochs and optimized it with AdamW \cite{loshchilov2017decoupled} optimizer with a 0.001 weight decay. The learning rate was initially warmed up to $1e^{-4}$ and then linearly decayed to $1e^{-5}$. For variable-length diverse-resolution video input, we employed the sparsely head-tail sampling strategy to derive $4*224*224$ video clips. During the fine-tuning stage, we raise the frame resolution to $384*384$ and tweak the sparse sample number to 8. To accommodate different scales and domains, we employed dataset-specific training epochs and learning rates based on validation results. During testing, we observe the standard test split settings and assess the quality of generated answers with the top-1 Accuracy \cite{xu2017video} and the Wu-Palmer similar (WUPS) \cite{malinowski2014multi} evaluation criteria. 

\subsection{State-of-the-Art Comparison}
We evaluate the performance of our MCG model by comparing it against the state-of-the-art methods on four publicly available VideoQA benchmarks. Detailed experimental comparisons and analyses are presented below.
\subsubsection{Comparison on AcitivityNet-QA}
In Table \ref{tab:com_ACT}, we present a comprehensive comparison of the MCG model with existing state-of-the-art models, including RNN-based approaches like EVQA \cite{zeng2017leveraging}, memory networks such as CoMem \cite{gao2018motion}, HME \cite{fan2019heterogeneous}, and MHMAN \cite{yu2020long}, as well as modular attention networks like AMU \cite{xu2017video} and CAN \cite{yu2019can}, all of which follow a two-stage paradigm. Additionally, we reproduce the typical video-language pre-training model ALPRO \cite{li2022align} on long-term videos as the representative pre-trained comparison model. Our experimental results yield several noteworthy findings:
ALPRO consistently outperforms all existing frameworks and achieves a remarkable 2.7\% improvement in overall accuracy over the state-of-the-art two-stage network MHMAN. This underscores the crucial role of external implicit knowledge acquired through pre-training in enhancing long-term video understanding and QA reasoning.
However, as impressive as ALPRO's performance is, it falls short of our proposed MCG model. MCG achieves state-of-the-art results across all evaluation metrics, surpassing ALPRO by 3.5\% in overall accuracy, 4.6\% in WUPS@0.9, and 3.8\% in WUPS@0.0. These results underscore the significance of end-to-end video rationale coupled with answer generation.
A closer examination reveals that MCG exhibits a remarkable 5.3\% improvement in motion-related tasks, a 4.8\% improvement in tasks involving spatial relationships, and a 5.2\% improvement in tasks related to temporal relationships when compared to the pre-trained SoTA model \cite{li2022align}. This notable performance enhancement in these critical aspects demonstrates MCG's ability to grasp both long video dependencies and spatial cues, which are particularly valuable in long-term VideoQA.
This detailed comparison on the ActivityNet-QA benchmark underscores the superiority of the MCG model in handling complex video understanding and reasoning tasks.
\begin{table*}	
	\centering
	\caption{Experimental comparative results of the proposed MCG and existing SOTAs on ActivityNet-QA benchmark. All records are listed as a percentage (\%). ``Spat. Rel." and ``Temp. Rel." abbreviate Spatial Relationship and Temporal Relationship tasks, respectively. ``Obj.," ``Loc.," and ``Num." correspond to Object, Location, and Number tasks, respectively. The symbol $\dag$ denotes a reproduction version implemented on long-term videos, and the best results are highlighted in bold.}
	\label{tab:com_ACT}
    \setlength{\tabcolsep}{1.9mm}
	\begin{tabular}{l|ccc|ccccccc|c|cc}
		\toprule
		\multirow{2}{*}{{Method}}&
		\multicolumn{11}{c}{{Accuracy (\%)}}&\multicolumn{2}{|c}{ {WUPS (\%)}}\\
        \cmidrule{2-4}\cmidrule{5-12}\cmidrule{13-14}
		&{Motion} & {Spat. Rel.} & {Temp. Rel.} & {Y/N} & {Color} & {Obj.} & {Loc.} & {Num.} & {Other} & {Free} & {All}& {@0.9}& {@0.0}\\
\midrule
		EVQA\cite{zeng2017leveraging} &2.5 & 6.6 & 1.4 &52.7&27.3&7.9&8.8&44.2&20.6&34.3&25.1& 29.3 & 53.5  \\
      CoMem\cite{gao2018motion} &16.1&13.8&2.9&58.3&29.8&15.4&18.1&44.4&29.3&40.3&31.5&34.3&56.1 \\
      AMU\cite{xu2017video}  &9.8&14.0&2.3&61.1&27.0&19.8&23.1&45.2&30.1&41.9&31.9&36.1&57.3 \\
	   CAN\cite{yu2019can}  &21.1&17.3&3.6&62.6&31.1&20.1&30.6&48.0&33.3&44.5&35.4&40.5&60.0 \\        
		HME\cite{fan2019heterogeneous} &19.4&14.1&3.0&59.7&31.6&16&19.4&45.4&30.0&41.4&32.7&36.7&57.1 \\
      MHMAN\cite{yu2020long} &23.1&19.8&4.4&63.2&32.1&22.0&33.2&48.5&36.8&46.2&37.1&42.8&62.1 \\
\midrule
ALPRO\dag \cite{li2022align}&23.5&19.3&2.9&71.3&36.4&25.5&35.0&52.5&35.7&50.3&39.8&43.9&65.3\cr
MCG &\textbf{28.8}&\textbf{25.1}&\textbf{8.1}&\textbf{73.1}&\textbf{38.9}&\textbf{30.8}&\textbf{38.9}&\textbf{55.6}&\textbf{38.7}&\textbf{53.0}&\textbf{43.3}&\textbf{48.5}&\textbf{69.1}\cr
		\bottomrule
	\end{tabular}
\end{table*}

\begin{table*}
	\centering
		\caption{Experimental comparative results of the proposed MCG and existing SOTAs on MSRVTT-QA (Left) and MSVD-QA (Right) benchmarks. All values are reported as percentages (\%). The symbol $\dag$ denotes a reproduction version conducted by us. The best outcomes are emphasized in bold.}
		\label{tab:msvd_msrvtt}
        \setlength{\tabcolsep}{2.5mm}

		\begin{tabular}{l|cccccc|cccccc}
			\toprule
			\multirow{3}{*}{Method}&
			\multicolumn{6}{c|}{MSRVTT-QA}&\multicolumn{6}{c}{ MSVD-QA}\cr
			\cmidrule(lr){2-7} \cmidrule(lr){8-13}
			&What&Who&How&When&Where&All&What&Who&How&When&Where&All\cr
			\midrule
			EVQA \cite{zeng2017leveraging}&18.9&38.7&83.5&70.5&29.2&26.4&9.7&42.4&83.8&72.4&53.6&23.3\cr
			STVQA \cite{jang2019video}&24.5&41.2&78.0&76.5&34.9&30.9&18.1&50.0&83.8&72.4&28.6&31.3\cr
         CoMem \cite{gao2018motion}&23.9&42.5&74.1&69.0&42.9&32.0&19.6&48.7&81.6&74.1&31.7&31.7\cr
			AMU \cite{xu2017video}&26.2&43.0&80.2&72.5&30.0&32.5&20.6&47.5&83.5&72.4&53.6&32.0\cr		 
          CAN \cite{yu2019compositional}&26.7&43.4&83.7&75.3&35.2&33.2&21.1&47.9&\textbf{84.1}&74.1&57.1&32.4\cr
          TSN \cite{yang2019question}&27.9&46.1&84.1&77.8&37.6&35.4&25.0&51.3&83.8&\textbf{78.4}&\textbf{59.1}&36.7\cr
          HME \cite{fan2019heterogeneous}&26.5&43.6&82.4&76.0&28.6&33.0&22.4&50.1&73.0&70.7&42.9&33.7\cr
          MHMAN \cite{yu2020long}&28.7&47.1&85.1&77.1&35.2&35.6&23.3&50.7&\textbf{84.1}&72.4&53.6&34.6\cr
          HGA \cite{jiang2020reasoning}&29.2&45.7&83.5&75.2&34.0&35.5&23.5&50.4&83.0&72.4&46.4&34.7\cr
\midrule         
ALPRO\dag \cite{li2022align}&36.0&51.7&\textbf{85.7}&79.6&42.3&41.9&36.0&57.0&82.2&72.4&46.6&44.7\cr
MCG&\textbf{38.1}&\textbf{53.9}&\textbf{85.7}&\textbf{81.2}&\textbf{45.2}&\textbf{44.0}&\textbf{39.4}&\textbf{60.6}&\textbf{84.1}&77.6&\textbf{57.1}&\textbf{48.2}\cr
			\bottomrule
		\end{tabular}
\end{table*}

\begin{table}
	\centering
		\caption{Experimental comparative results of accuracy on NExT-QA dataset. All records are listed as a percentage (\%). The ``C'', ``T'', and ``D'' refer to the Causal, Temporal, and Descriptive tasks, respectively. The best records are emphasized in bold.}
		\label{tab:nextqa_mulchioce}
       \setlength{\tabcolsep}{2.5mm}
		\begin{tabular}{l|cccc}
			\toprule
			\multirow{3}{*}{Method}&
			\multicolumn{4}{c}{NExT-QA}\cr
			\cmidrule(lr){2-5}
				&Acc@C&Acc@T&Acc@D&Acc@A\cr
			\midrule
			  EVQA \cite{zeng2017leveraging}&43.27&46.93&45.62&44.92\cr
			  STVQA \cite{jang2019video}&45.51&47.57&54.59&47.64\cr 
            CoMem \cite{gao2018motion}&45.85&50.02&54.38&48.54\cr
            HCRN \cite{le2020hierarchical}&47.07&49.27&54.02&48.89\cr
            HME \cite{fan2019heterogeneous}&46.76&48.89&57.37&49.16\cr          
            HGA \cite{jiang2020reasoning}&48.13&49.08&57.79&50.01\cr
            IGV \cite{li2022invariant}&48.56&51.67&59.64&51.34\cr
         \midrule
           JustAsk \cite{yang2021just}&41.66&44.11&59.97&45.30\cr
           ATP \cite{buch2022revisiting}&53.10&50.20&66.80&54.30\cr
           VGT \cite{xiao2022video}&52.78&54.54&67.26&55.70\cr
         \midrule
           MCG&\textbf{57.31}&\textbf{56.45}&\textbf{68.19}&\textbf{58.83}\cr
			\bottomrule
		\end{tabular}
\end{table}
\subsubsection{Comparison on MSRVTT-QA and MSVD-QA}
We further investigate the performance of the MCG model on short-term videos by evaluating it on the MSRVTT-QA and MSVD-QA datasets. The results of this comparison are presented in Table \ref{tab:msvd_msrvtt}.
From the experimental results, it is evident that MCG exhibits a significant improvement over previous state-of-the-art models on both benchmarks. Notably, it achieves an impressive 8.4\% and 11.5\% increase in overall accuracy, demonstrating its effectiveness in handling short-term video comprehension tasks. While the performance improvements are relatively modest compared to long-term video datasets, these results underscore the remarkable capabilities of the entirely end-to-end MCG structure.
To delve deeper into the reasons behind this achievement, we reproduce the representative VLP model ALPRO, which leverages external pre-training on 5.5 million video-text pairs. As expected, ALPRO substantially enhances performance, achieving 41.9\% overall accuracy on MSRVTT-QA and 44.7\% overall accuracy on MSVD-QA. This surpasses the hierarchical graph-based network HGA \cite{jiang2020reasoning}, equipped with sophisticated reasoning mechanisms, by a margin of 6.4\% and 10.0\% in accuracy.
These findings highlight the value of external knowledge involving weak supervision pre-training is conducive to delivering a good performance on video question answering tasks. Surprisingly, our MCG model surpasses this newly pre-trained SoTA model by 2.1\% on MSRVTT-QA and 3.5\% on MSVD-QA. Moreover, when considering specific question types, MCG consistently achieves the highest accuracy on both MSRVTT-QA and MSVD-QA, illustrating its competence in short-term video descriptive analysis.

\subsubsection{Comparison on NExT-QA}
We conducted additional experiments on the challenging NExT-QA dataset, which places a strong emphasis on temporal and causal reasoning in a multi-choice task. To ensure a fair comparison, we adjusted the MCG model to adapt to the multi-choice paradigm by calculating the video-text matching score for each video and the answer candidates.
As depicted in Table \ref{tab:nextqa_mulchioce}, the proposed MCG outperforms prior state-of-the-art models across all sub-questions, including causal, temporal, and descriptive tasks. Specifically, compared to the two-stage SoTA model IGV \cite{li2022invariant}, which is conditioned on invariant grounding without cross-modal pre-training, MCG achieves a remarkable 7.4\% increase in overall accuracy and an incredible 8.7\% improvement in the causal reasoning task. This improvement is attributed to the collaborative reasoning capability of MCG.
We also conducted a detailed examination of prior SoTA models, namely VLP JustAsk \cite{yang2021just}, ATP \cite{buch2022revisiting}, and VGT \cite{xiao2022video}, which are pre-trained on large-scale external web-sourced data. The experimental results are presented in the second split of Table \ref{tab:nextqa_mulchioce}. Interestingly, we observe that these typically pre-trained models on NExT-QA have seen limited performance boosting in temporal and causal reasoning types. This limitation may be attributed to the domain gap between externally sourced data and QA-specific data. However, MCG, which incorporates both external and domain-specific data, effectively mitigates domain disconnection issues and consistently outperforms these competitive video-language pretraining counterparts \cite{yang2021just, buch2022revisiting, xiao2022video}. It achieves absolute overall accuracy gains of 13.5\%, 4.5\%, and 3.3\%, respectively, highlighting its effectiveness and causal reasoning advantage.
These results underscore the superior performance and causal reasoning capabilities of the MCG model on the NExT-QA dataset.

\subsection{Ablation Study}
In this section, we conduct a comprehensive ablation analysis on the ActivityNet-QA dataset. Our objective is to investigate the effects of several critical components: joint unimodal modeling, multi-granularity contrastive learning, and the cross-modal collaborative generation module in the context of long-term videos. Additionally, we delve into the influence of varying video sampling rates for variable-length videos across different benchmarks, explore the impact of model parameters and pre-training data size, and assess MCG's generalization ability across different video scenarios.
\begin{table}	
	\centering
	\caption{Ablation studies on the ActivityNet-QA test split. All values are presented as percentages (\%). ``Spa. R." and ``Tem. R." abbreviate spatial relationship and temporal relationship, respectively.}
	\label{tab:aba}
	\begin{tabular}{c|ccccc}
\toprule
	Methods&Motion&Spa.R.&Tem.R.&Free&All \\
\midrule
     Baseline &18.8&14.4&3.0&43.8&34.3\cr
	w/ JUM&24.4&21.9&3.5&47.3&38.2\cr
     w/ JUM\&MCL &25.1&21.4&4.3&51.1&41.4\cr
     w/ JUM\&MCL\&CCG &28.8&25.1&8.1&53.0&43.3\cr
\midrule
     MCG w/o TCL&23.8&20.0&4.3&49.9&39.8\cr
     MCG w/o ICL&25.0&23.8&4.4&52.3&41.9\cr
     MCG w/ ICL\&TCL&28.8&25.1&8.1&53.0&43.3\cr
\midrule
     MCG-CH&25.1&21.4&4.3&51.1&41.4\cr
     MCG&28.8&25.1&8.1&53.0&43.3\cr
		\bottomrule
\end{tabular}
\end{table}

\begin{table}	
	\centering
	\caption{Ablation studies investigating the impact of different video frame sampling rates on four video question answering benchmarks. All records are reported as a percentage (\%).}
	\label{tab:frames}
	\begin{tabular}{c|ccccc}
	\toprule
		\#Frames&2&4&8&12&16 \\
	\midrule
     ActivityNet-QA&39.8&41.6&43.3&44.1&44.3\cr
     NExT-QA&54.3&56.9&58.8&59.6&59.9\cr
     MSRVTT-QA&42.1&42.9&44.0&44.2&44.3\cr
     MSVD-QA&46.3&47.1&48.2&48.3&48.3\cr
	\bottomrule
\end{tabular}
\end{table}
\begin{figure}
	\centering
	\includegraphics[width=0.48\textwidth]{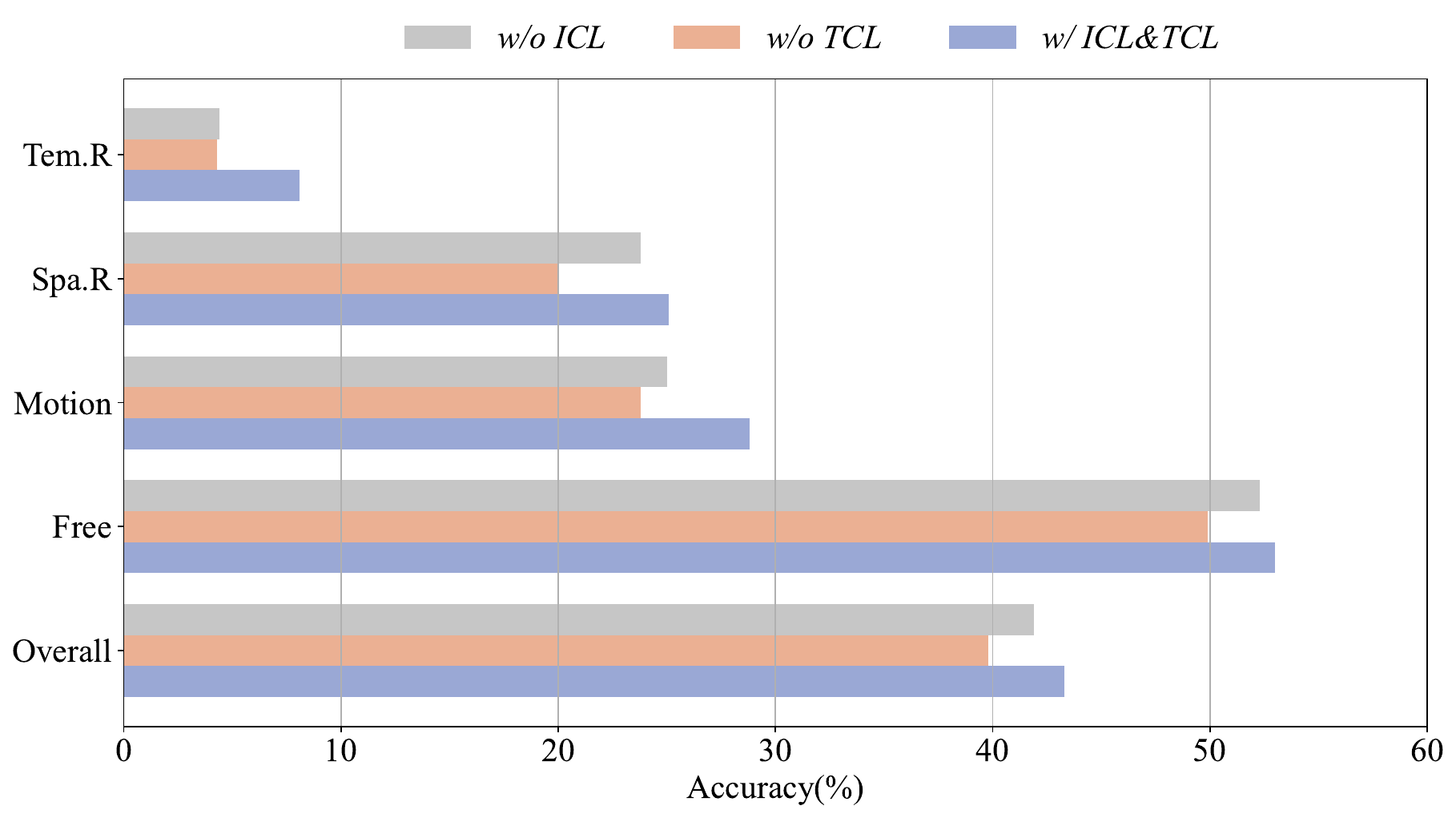}
	\caption{
	In-depth ablation study comparing multi-granularity contrastive learning (w/ ICL\&TCL) with uni-granularity contrastive learning (w/o ICL and w/o TCL) across various task types, including temporal relationship (Tem. R.), spatial relationship (Spa. R.), motion, and free tasks. Best viewed in color.}
	\label{fig:aba_mcl}
\end{figure}

\begin{figure}
	\centering
	\subfigure[MCG-CH] {\includegraphics[width=0.49\linewidth]{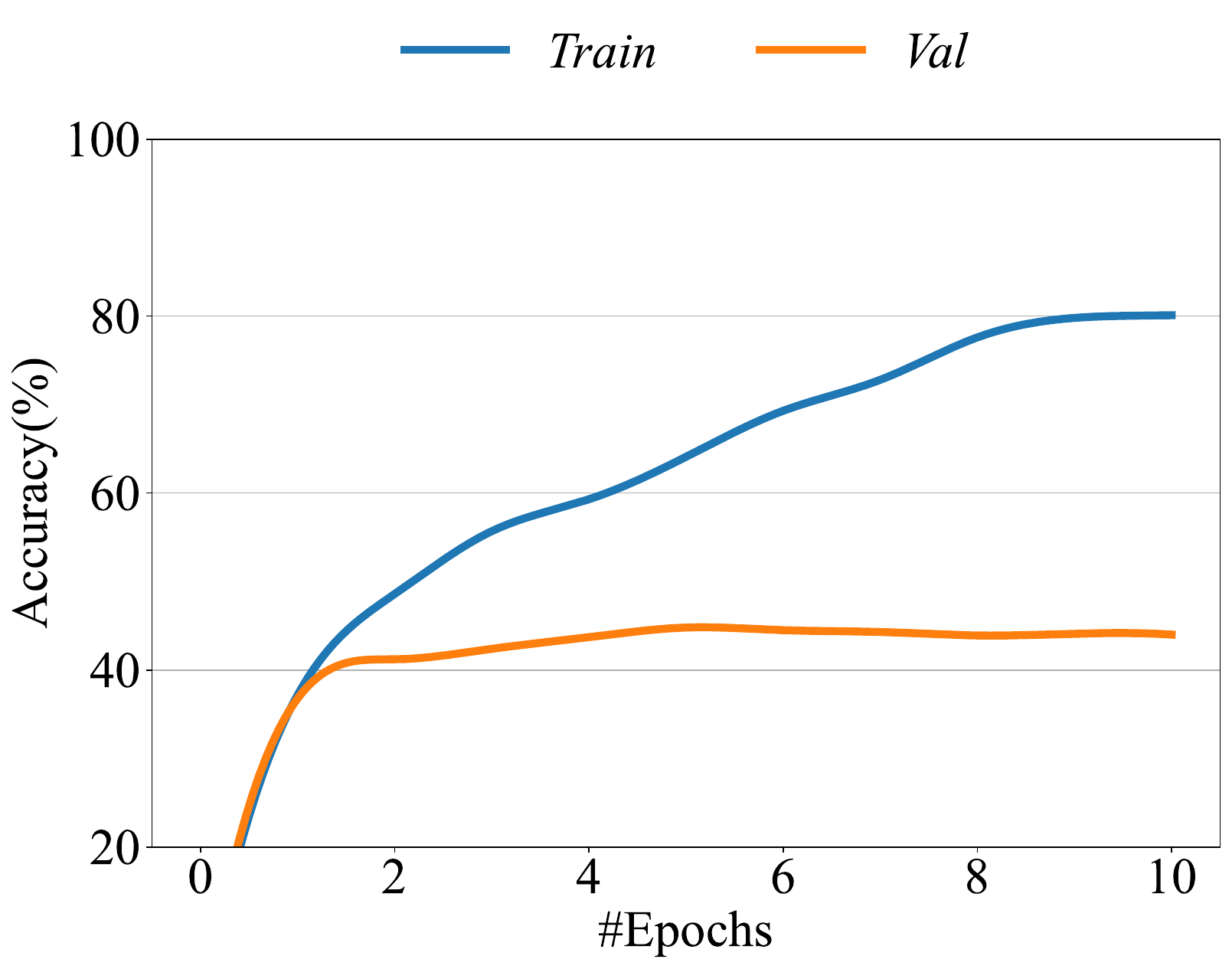}\label{fig:acc_v}}
	\subfigure[MCG] {\includegraphics[width=0.49\linewidth]{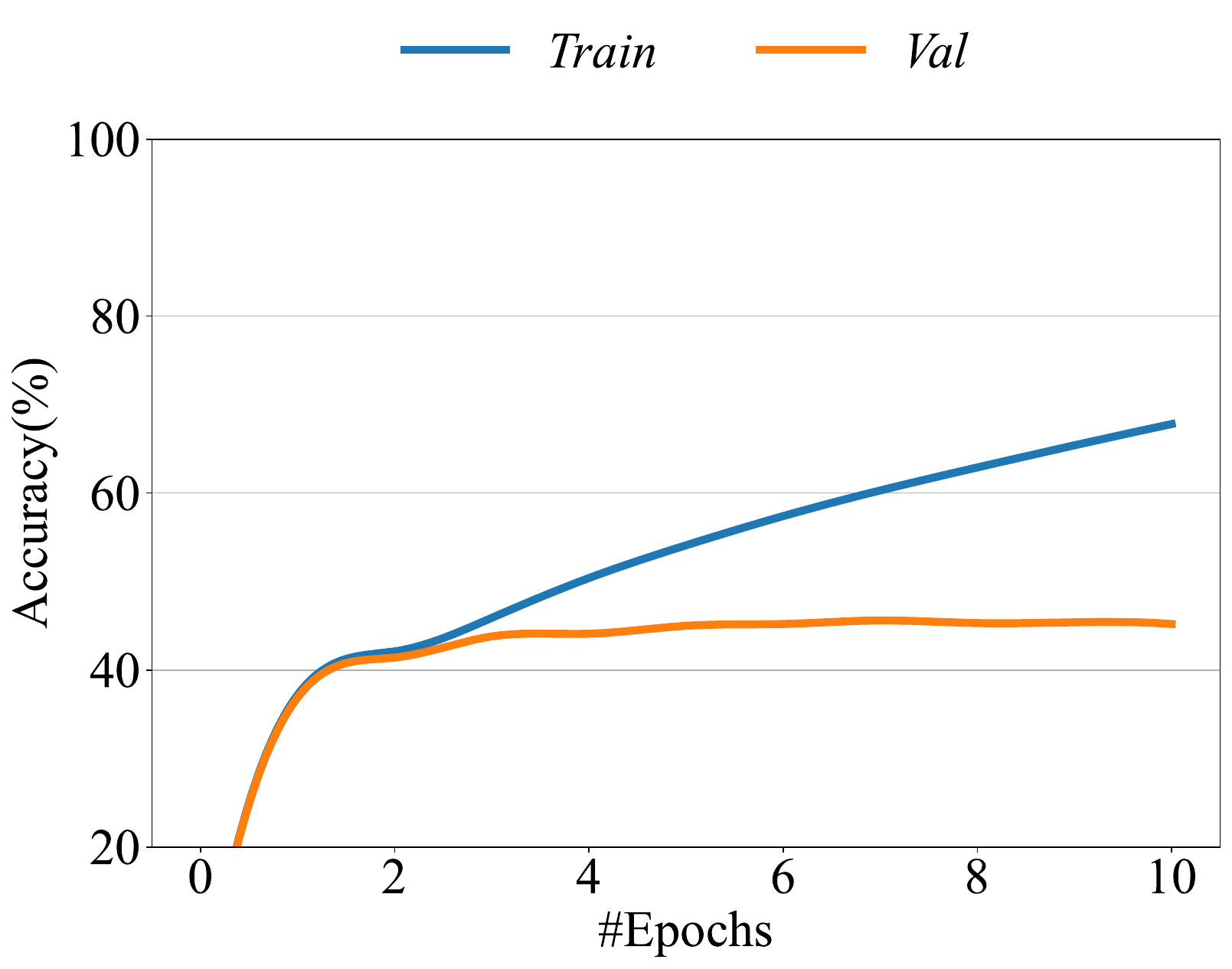}\label{fig:wups_v}}	
	\caption{
	Visualization of training and validation accuracy trends across various training epochs for (a) MCG-CH, a model variant employing a Classifier Head, and (b) MCG, the proposed model with language generation capability. Optimal viewing experience in color.}
	\label{fig:aba_overfiting}
\end{figure}
\subsubsection{MCG vs. Baseline}
In the top split of Table \ref{tab:aba}, we aim to assess the individual contributions of various components within the MCG model. To establish a baseline, we systematically remove or replace key components from the complete MCG system. These components include the JUM module, the MCL strategy, and the CCG module substituted with the AGor. Specifically, we replace the JUM with offline feature extractors \cite{yu2019compositional}, discard the MCL strategy, and replace the AGor in the CCG with a classification head in the system baseline.
\begin{figure*}
	\centering
	\subfigure[AcitivityNet-QA] {\includegraphics[width=0.24\linewidth]{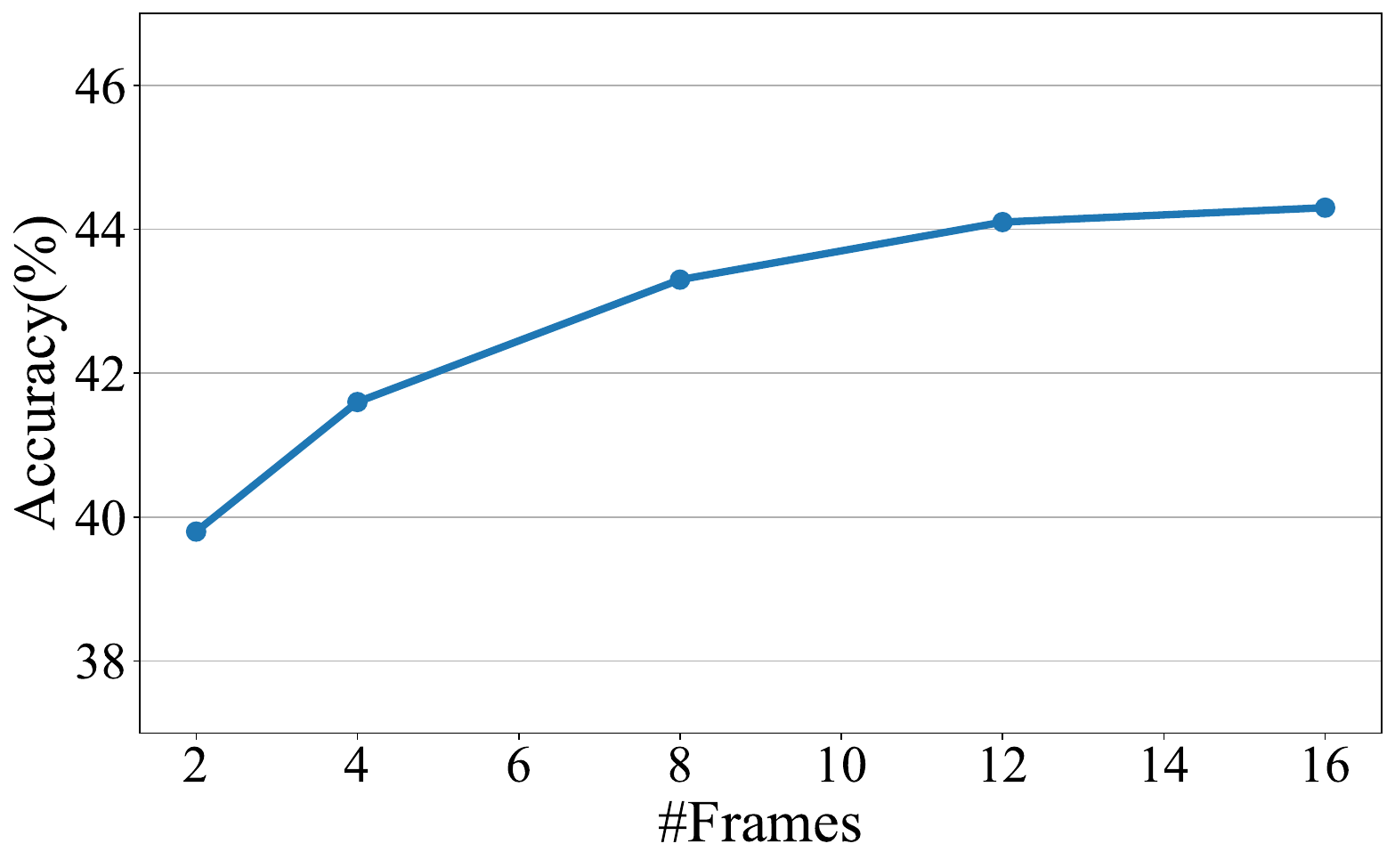}\label{fig:act}}
	\subfigure[NExT-QA] {\includegraphics[width=0.24\linewidth]{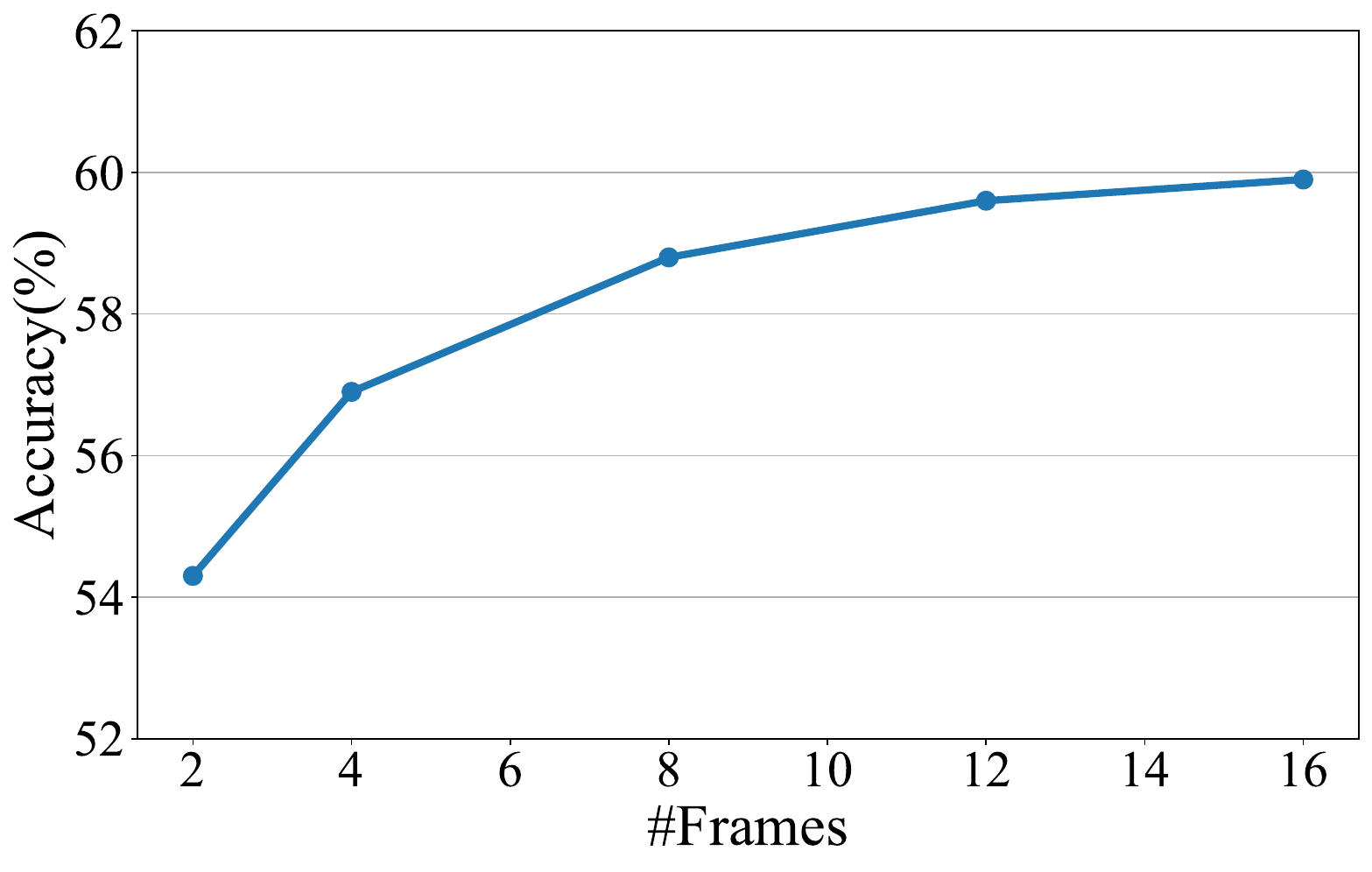}\label{fig:nex}}
	\subfigure[MSRVTT-QA] {\includegraphics[width=0.24\linewidth]{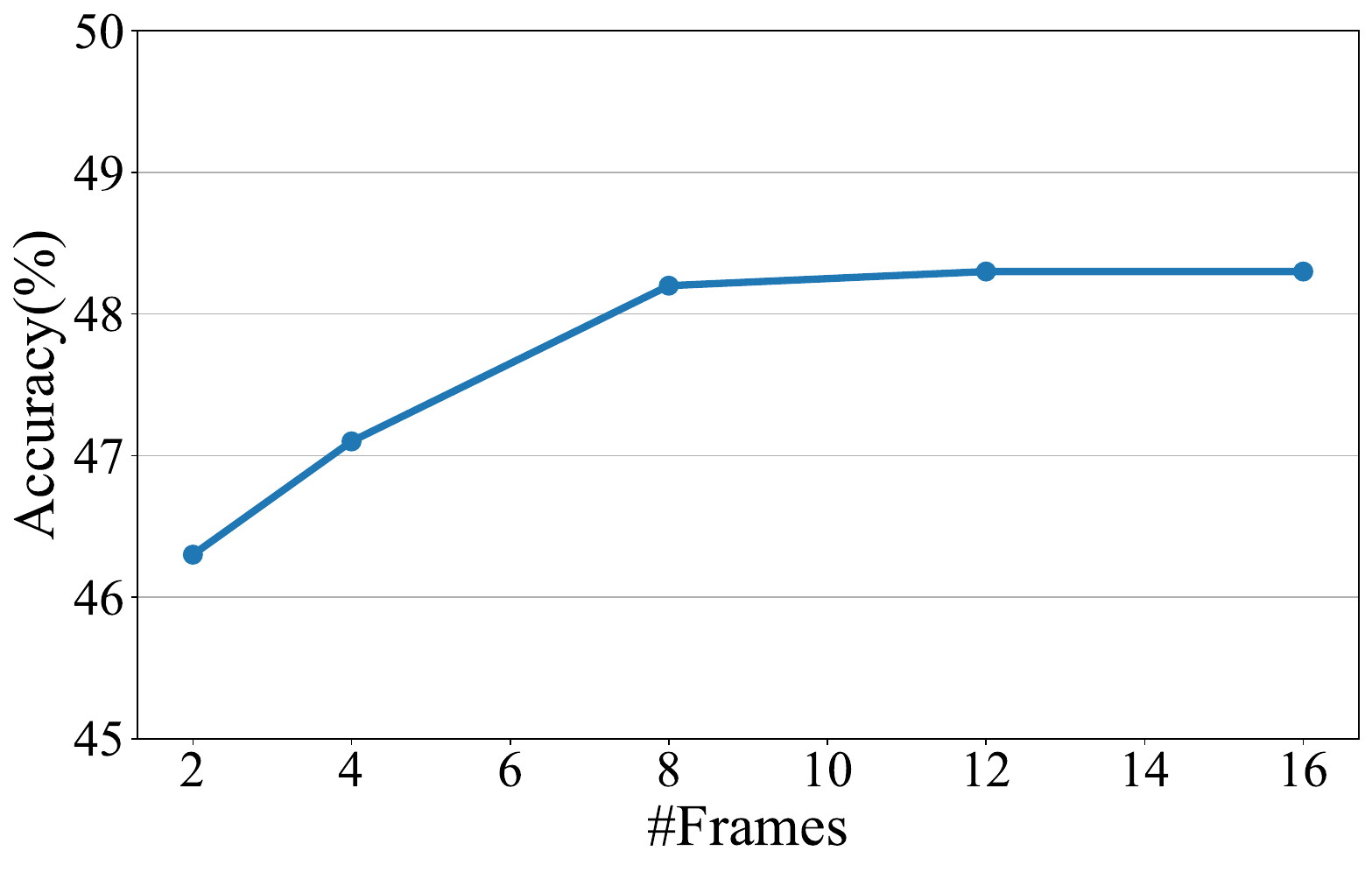}\label{fig:msr}}
   \subfigure[MSVD-QA] {\includegraphics[width=0.24\linewidth]{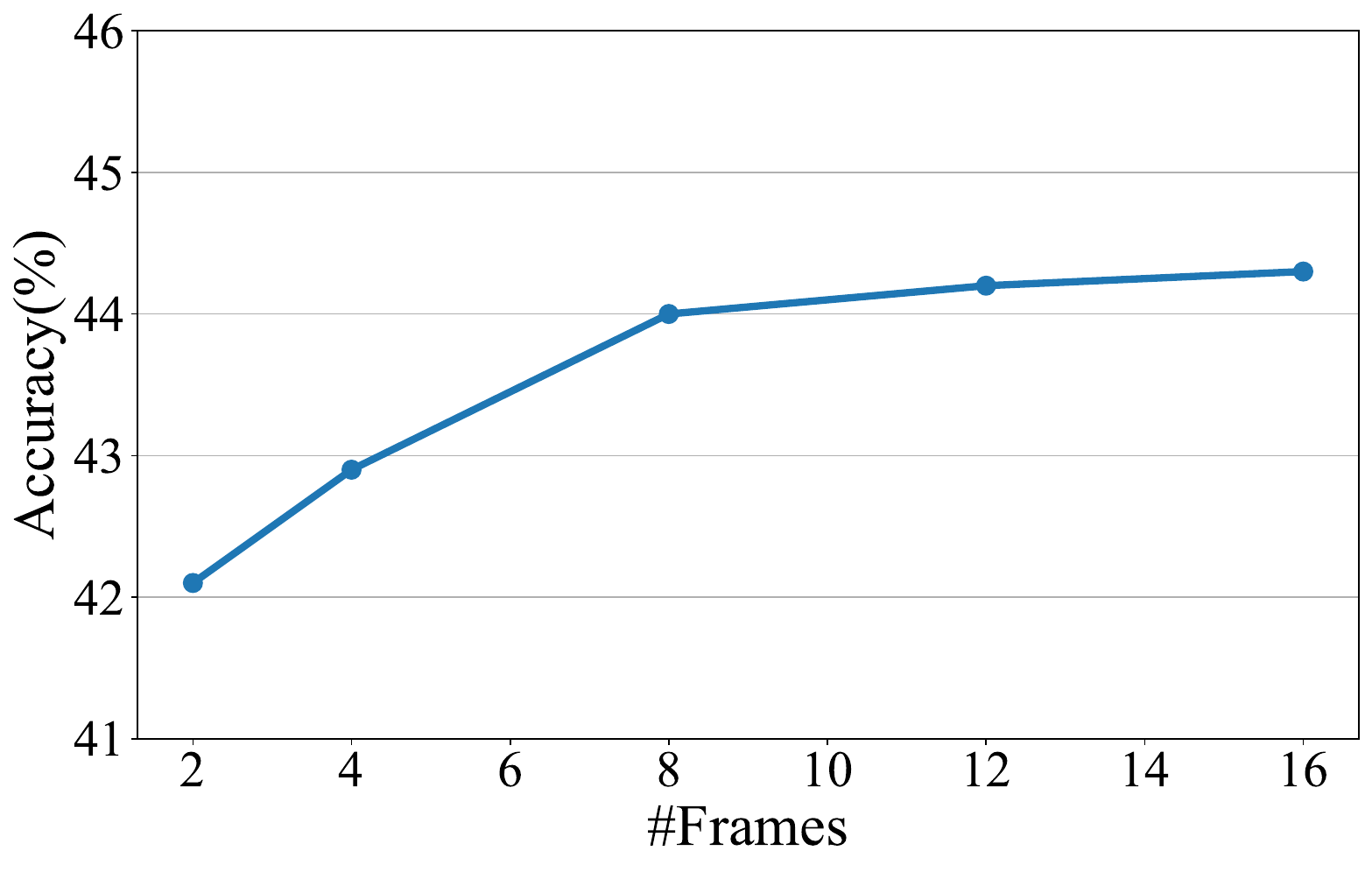}\label{fig:msv}}
	\caption{Variations in test accuracy by employing different video frame sampling rates on (a) ActivityNet-QA, (b) NExT-QA, (c) MSRVTT-QA, and (d) MSVD-QA. The number of sampling frames ranges from 2 to 16. A consistent performance boost is observed when the number of samples varied from 2 to 8 while gradually reaching saturation from 8 to 16. Best viewed in color.}
	\label{fig:aba_frames}
\end{figure*}
\textbf{JUM:} We begin by examining the impact of JUM, and the ablation results are presented in the first two rows of Table \ref{tab:aba}. When compared to the system baseline, equipping the model with joint unimodal modeling (w/ JUM) in clip fashion \cite{radford2021learning} leads to a consistent improvement in performance. Specifically, the overall accuracy increases by 3.9\%, with notable performance enhancements observed across all sub-questions. Notably, tasks related to motion, spatial relationships, and free-form tasks see substantial improvements, with scores increasing by 5.6\%, 6.5\%, and 3.5\%, respectively. These results underscore the effectiveness of joint unimodal modeling in expressing discriminative representations with the supervision of complementary modalities, thereby playing a vital role in enhancing video question-answering capabilities.

\begin{table}
	\centering
		\caption{Comparison with existing video-language pre-training models concerning model parameter scale and pre-training data size.}
		\label{tab:parameter}
		\begin{tabular}{lcc}
			\toprule
			{Method}&{PT Data}&Model Size\cr
         \midrule
           ClipBERT \cite{lei2021less}&COCO\cite{chen2015microsoft}\&VG\cite{krishna2017visual}(5.6M)&137M\cr
           JustAsk\cite{yang2021just}&HTVQA\cite{yang2021just}(69M)&600M\cr
           ATP \cite{buch2022revisiting}&WebIT\cite{radford2021learning}(400M)&428M\cr
           VGT \cite{xiao2022video}&WV\cite{bain2021frozen}(0.18M)&511M\cr
           ALPRO\cite{li2022align}&WV\cite{bain2021frozen}\&CC\cite{sharma2018conceptual}(5.5M)&240M\cr
           FrozenBiLM \cite{yang2022frozenbilm}& WebVid10M\cite{bain2021frozen}(10M)&890M\cr
           MCG& WV\cite{bain2021frozen}\&CC\cite{sharma2018conceptual}\&TGIF\cite{jang2017tgif}(5.6M)&297M\cr
	  \bottomrule
		\end{tabular}
\end{table}

\begin{table}
\centering
\caption{Comparison of direct image-language model adaptation to video question answering on ActivityNet-QA.}
\label{tab:blip_performance}
\setlength{\tabcolsep}{1.7mm}
\begin{tabular}{lcccc}
\toprule
Method & \#Total Params & \#Trainable Params & \#PT Data & Acc. \cr
\midrule
BLIP \cite{li2022blip}& 305M & 305M & 129M & 24.2 \cr
BLIP-2 \cite{li2023blip} & 3.1B & 104M & 129M & 35.4 \cr
MCG & 297M & 297M & 5.6M & 43.3 \cr
\bottomrule
\end{tabular}
\end{table}

\textbf{MCL:} To assess the performance of MCL, we enhance the baseline by incorporating MCL into the JUM module (w/ JUM\&MCL). The experimental results demonstrate a notable 7.1\% improvement in overall accuracy when MCL is combined with JUM. This outcome validates our hypothesis that MCL has the capability to harness hierarchical intrinsic semantic correspondences, thereby facilitating cross-modality understanding.

\textbf{CCG:} We further explore the impact of the CCG module. When we add the collaborative generation module to the baseline system equipped with JUM and MCL, denoted as w/ JUM\&MCL\&CCG, our model experiences significant performance improvement, achieving the best performance. Quantitatively, the CCG module contributes to an average 3.7\% gain in motion-related, spatial relationship-related, and temporal relationship-related tasks, ultimately resulting in an overall accuracy of 43.3\%. These substantial enhancements underscore the effectiveness of the CCG module.

\begin{figure}
	\centering
	\includegraphics[width=0.49\textwidth]{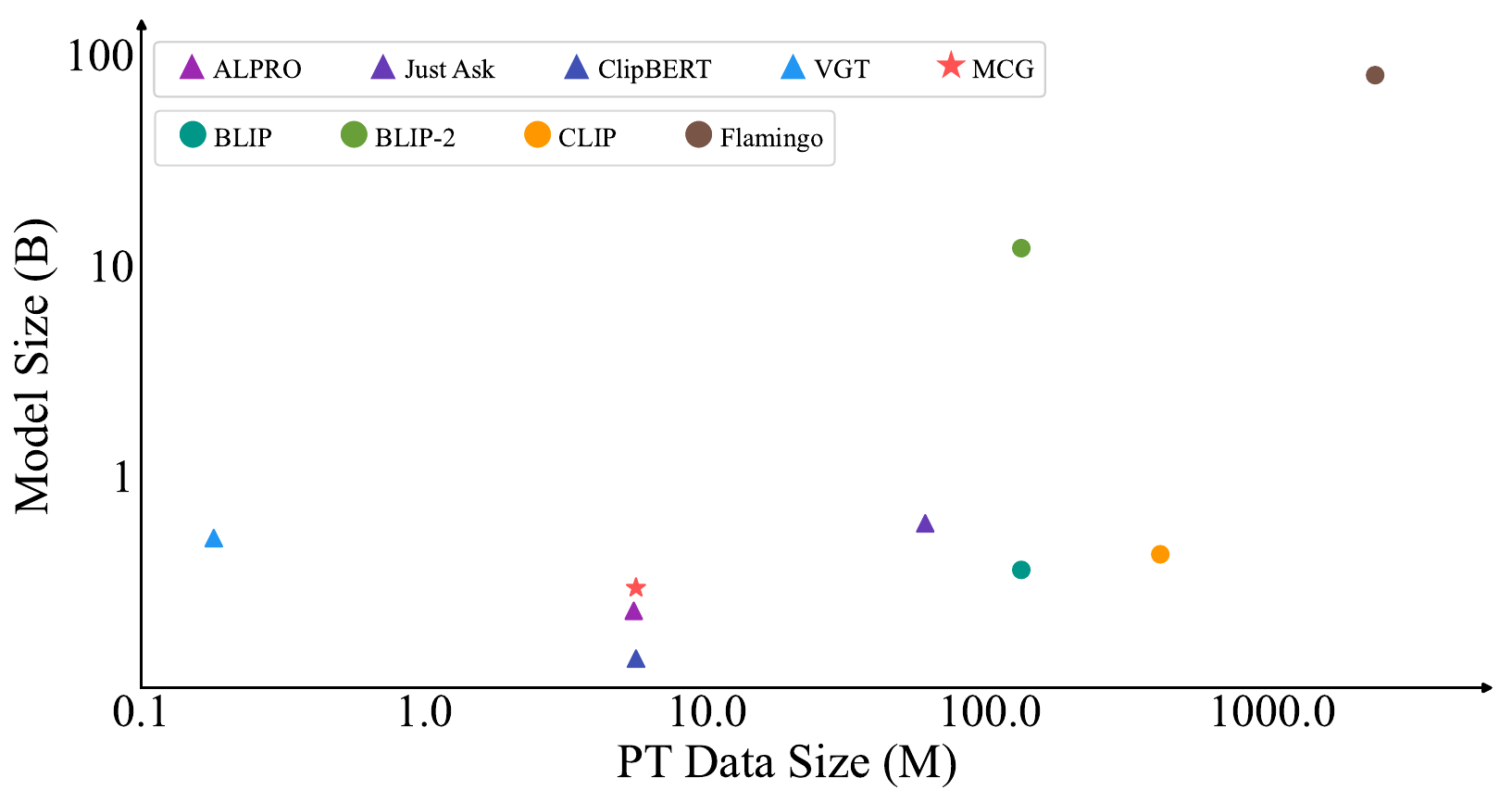}
	\caption{Comparison of Image-Language and Video-Language pre-training models concerning model parameter scale and pre-training data size.}
	\label{fig:param_data_statistic}
\end{figure}
\subsubsection{Multi-Granularity vs. Uni-Granularity}
In the middle part of Table \ref{tab:aba}, we conduct an in-depth analysis of MCL by comparing it to uni-granularity contrastive learning. Firstly, we remove the TCL loss from the overall training objectives, effectively setting $\theta_{2}$ to 0 in Eqn.\ref{eq:loss_mcl}, and retain only the uni-granularity ICL loss. This configuration is represented as MCG w/o TCL. The comparison results clearly demonstrate that eliminating the TCL loss leads to a noticeable drop in performance across all sub-tasks.
Similarly, we also experiment with removing the ICL loss from the full model by setting the hyperparameter $\theta_{1}$ to 0 in Eqn.\ref{eq:loss_mcl}, referred to as MCG w/o ICL. Once again, the results show significant performance degradation compared to the multi-granularity contrastive model. Detailed comparisons are visualized in Figure \ref{fig:aba_mcl}. In summary, both uni-granularity contrastive losses (i.e., ICL or TCL) play a crucial role in performance enhancement, while multi-granularity contrastive learning, which incorporates both of them, achieves the best performance.

\subsubsection{Generation vs. Classification}
We study MCG's variant by replacing the answer generator with a classifier head. Specifically, the output of the $[$FUS$]$ signal from the cross-modal fusion module is directed to a K-way classifier following \cite{yu2021learning,lei2021less,li2022align}. As illustrated in the bottom block of Table \ref{tab:aba}, this classification variant model, referred to as MCG-CH, results in a significant drop in performance, highlighting that the task-specific classifier head lacks the depth of reasoning required for video question answering. To further verify our assumption that incorporating a task-disparity classifier head inevitably brings in unexpected extra parameters and may lead to severe over-fitting in video question answering, we conducted in-depth experiments to analyze the performance on both training and validation data, respectively. Figure \ref{fig:aba_overfiting} provides a visual representation of training and validation scores across various training epochs. The experimental results indeed confirm the presence of overfitting in the classifier variant model MCG-CH. Moreover, it's evident from the trends in the MCG curves that MCG with a generative solution can mitigate the overfitting problem to some extent. In summary, the ablation results underscore the superiority of solving question-answering tasks in a generative paradigm and simultaneously subtly alleviate the task formulation discrepancy.
\begin{table}
	\centering
		\caption{Experimental comparative results of the proposed MCG and existing SOTAs on TVQA. Overall accuracy is listed as a percentage (\%). Models that take extra multi-modal television inputs are grayed out. The best outcomes are emphasized in bold.}
		\label{tab:tvqa}
       \setlength{\tabcolsep}{5.5mm}
	\begin{tabular}{lc}
	\toprule
	{Method}&{Accuracy}\cr
         \midrule
     \multicolumn{2}{l}{{\color{gray}\emph{w/ multi-modal television inputs}}}\cr
     {\color{gray}Human \cite{lei2018tvqa} }& {\color{gray}89.4}\cr
   {\color{gray}HCRN \cite{le2020hierarchical}}& {\color{gray}71.3}\cr
   {\color{gray}HERO \cite{li2020hero}}& {\color{gray}73.6}\cr
   {\color{gray}FrozenBiLM \cite{yang2022frozenbilm}}&{\color{gray}82.0}\cr
           \midrule
            \multicolumn{2}{l}{\emph{w/ only visual inputs}}\cr
          {Human \cite{lei2018tvqa}} &61.9\cr
            ALPRO\dag \cite{li2022align}&57.2\cr
            FrozenBiLM \cite{yang2022frozenbilm} w/o speech &57.5\cr
           MCG& \textbf{59.8}\cr
	  \bottomrule
		\end{tabular}
\end{table}
\subsubsection{Sparse Sampling vs. Dense Sampling}
In Table \ref{tab:frames}, we investigate the influence of video frames sampling rate across all four video question answering benchmarks. Intuitively, increasing the number of frames should bring in more information and potentially facilitate the delivery of better performance. Figure \ref{fig:aba_frames} illustrates the performance trends of the model as the number of sampled frames varies from sparse to dense. From the evaluation curve trends, we observe that the model consistently experiences rapid performance improvement across all benchmarks when using sparse samples, ranging from 2 to 8 frames. However, as the number of sampled frames gradually increases from 8 to 16, the performance tends to reach saturation, especially for short-term videos. Even for the ActivityNetQA dataset, which features prolonged videos, the model's performance approaches saturation when the number of samples reaches eight, with only slight accuracy gains beyond this point. Considering the balance between computational efficiency and model performance, we have consistently set the sampling number to 8 for our final MCG across all benchmarks. Sparse sampling not only enables efficient end-to-end modeling but also benefits video question answering with high quality and efficiency.
\begin{table}	
	\centering
	\caption{Experimental comparative results of the proposed MCG and existing SOTAs on CLEVRER benchmark. All records are listed as a percentage (\%). Descrip.,  Expla., Predict., and Count. represent Descriptive, Explanatory, Predictive, and Counterfactual tasks, respectively.  Neural-symbolic and object-centric methods are grayed out.}
	\label{tab:com_CLV}
	\begin{tabular}{l|cccc|c}
		\toprule
		{{Method}}&{{Descrip.}}&{{Expla.}}&{{Predict. }}&{ {Count.}}&{{Overall}}\\
\midrule
 \multicolumn{6}{l}{{\color{gray}\emph{Neural-Symbolic or Object-Centric Methods}}}\cr
 {\color{gray}IEP(V) \cite{johnson2017inferring}}&{\color{gray}52.8}&{\color{gray}14.5}&{\color{gray}9.7}&{\color{gray}3.8}&{\color{gray}20.2}\\
{\color{gray}NS-DR \cite{CLEVRER2020ICLR}}&{\color{gray}88.1}&{\color{gray}79.6}&{\color{gray}68.7}&{\color{gray}42.4}&{\color{gray}69.7}\\
{\color{gray}DCL-O \cite{chen2021grounding}}&{\color{gray}91.4}&{\color{gray}82.0}&{\color{gray}82.1}&{\color{gray}46.9}&{\color{gray}75.6}\\
{\color{gray}ALOE \cite{ding2021attention}}&{\color{gray}94.0}&{\color{gray}96.0}&{\color{gray}87.5}&{\color{gray}75.6}&{\color{gray}88.3}\\
{\color{gray}VRDP \cite{ding2021dynamic}}&{\color{gray}93.4}&{\color{gray}91.9}&{\color{gray}91.4}&{\color{gray}84.3}&{\color{gray}90.3}\\
\midrule
Q-type &29.2&8.1&25.5&10.3&18.3\\
HME \cite{fan2019heterogeneous}&54.7&13.9&33.1&{7.0}&27.2\\
HCRN \cite{le2020hierarchical}&55.7&{21.0}&{21.0}&\textbf{11.5}&27.3\\
MCG&\textbf{61.6}&\textbf{34.2}&\textbf{49.2}&10.5&\textbf{38.8}\\
\bottomrule
\end{tabular}
\end{table}
\subsubsection{Model Parameters and Pre-training Data Size}
In Table \ref{tab:parameter}, we present a comprehensive comparison of MCG with other video-language pre-trained models, focusing on model size and pre-training data size. It is generally understood that larger models, supported by extensive pre-training data, typically yield enhanced performance. Remarkably, our MCG model demonstrates superior results when compared to models like ALPRO, which are similar in scale concerning both model size and pre-training data size. MCG even surpasses larger-scale architectures such as JustAsk, which has 600 million parameters and benefits from a large, domain-specific pre-training dataset. This demonstrates that MCG's performance is not merely a function of its parameter count or data scale.

We further extend our analysis to Image-Language pre-trained Models (ILMs) \cite{li2023blip, alayrac2022flamingo, li2022blip, radford2021learning} versus Video-Language pre-training models (VLMs) \cite{lei2021less, li2022align, xiao2022video, yang2021just}, as visualized in Figure \ref{fig:param_data_statistic}. 
Since the increased accessibility of image data and the comparatively simpler data structure, ILMs have advanced more rapidly than VLMs. As the figure illustrates, VLMs generally exhibit smaller scales in terms of model and pre-trained data sizes. This discrepancy arises from the cost and complexity associated with obtaining video-level annotations due to the temporal nature of videos and their richer content.
The notable gap between ILMs and VLMs highlights the urgent need for more efficient tuning methods and strategies for adapting powerful image-language models to VideoQA tasks.

To provide preliminary insights into the adaptability of image-language pre-trained models to video question answering, we conducted two sets of exploratory experiments. Table  \ref{tab:blip_performance} presents a comparative performance of BLIP ({ViT-B}) \cite{li2022blip}, BLIP-2 ({ViT-L-OPT2.7B}) \cite{li2023blip}, and our MCG model on ActivityNet-QA. To adapt BLIP to the video, we randomly selected a single frame per video to ensure a straightforward end-to-end training process without modifying the model's architecture. We adapted BLIP-2 to video with a concatenation setting, where the uniformly sampled eight frames were processed by the ViT-L \cite{radford2021learning} and the Q-former, with OPT \cite{opt} taking the concatenated visual features as the prefix. 
The results from these experiments have been insightful, revealing a performance gap due to domain differences and the lack of temporal modeling when directly adapting image-language pre-training models to VideoQA. Despite this, BLIP-2's performance improved significantly compared to BLIP, underscoring the potential of image-to-video transfer learning and parameter-efficient tuning techniques for complex video understanding.

\subsubsection{Generalization on TVQA}
Table \ref{tab:tvqa} evaluates  MCG's generalization performance in multi-modal television scenarios with the TVQA \cite{lei2018tvqa} dataset. To emphasize the model's understanding of inter-video elements and the reasoning behind their correlations with questions, MCG intentionally excludes the utilization of television subtitles or audio information, focusing solely on visual content to derive answers. 
To ensure a fair comparison, methods utilizing additional multi-modal television inputs are grayed out.
The experimental results reveal several key insights. First, the existing SoTA FrozenBiLM exhibits a significant 24.5\% performance gap between scenarios with and without speech input. This observation suggests that the model's comprehension is influenced by factors beyond visual content alone, possibly including cues from speech. When comparing MCG with the leading FrozenBiLM model, MCG exhibits a promising 2.6\% performance gain. This comparison signifies MCG's robustness in capturing intricate inter-modal relationships and reasoning within televisions.
Moreover, focusing solely on video visual information, MCG attains an impressive 59.8\% accuracy, closely approaching the human accuracy of 61.9\%. This experimental achievement illuminates MCG's certain ability to generalize across film and television scenarios.
\begin{figure*}
	\centering
	\includegraphics[width=0.9\textwidth]{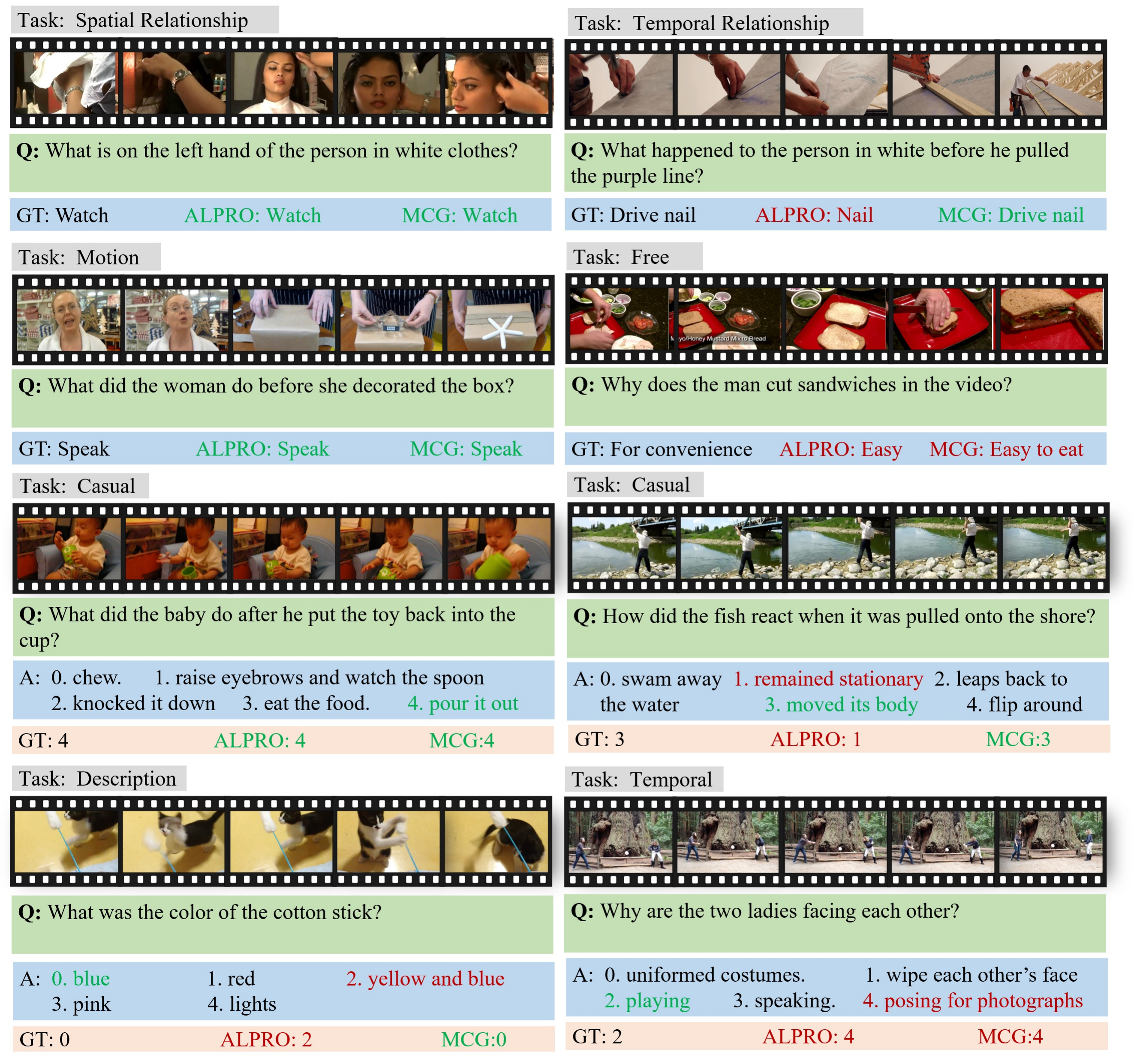}
	\caption{Visulization of question answering results from the proposed MCG for both open-end tasks and multi-choice tasks. The top two rows display four examples from ActivityNet-QA targeting open-end QA, including spatial relationship, temporal relationship, motion, and free tasks. The bottom half illustrates the multi-choice instances from NExT-QA, covering casual, temporal, and description tasks. Correct answers and Ground Truth (GT) are highlighted in green, while incorrect responses are marked in red.}
	\label{fig:vis_exam}
\end{figure*}
\subsubsection{Generalization on CLEVRER}
Table \ref{tab:com_CLV} investigates  MCG's generalization ability in synthetic object-oriented videos with the diagnostic CLEVRER \cite{CLEVRER2020ICLR} dataset. We gray out object-centric or neural-symbolic reasoning approaches that incorporate symbolic program execution bridging video concept acquisition and language semantic parsing. We observed that MCG outperforms existing VideoQA methods, demonstrating a notable 11.5\% increase in overall per-task accuracy. In open-ended descriptive tasks, MCG shows a significant 5.9\% improvement, underscoring its ability to interpret video content and engage in temporal reasoning. Additionally, MCG displays strong performance in explanatory and predictive tasks with improvements of 13.2\% and 16.1\%, respectively, indicating its certain ability in causal inference and future event anticipation. 

Yet, when compared with the ALOE  \cite{ding2021attention} model, MCG's limitations become apparent, particularly in the counterfactual domain where ALOE's discrete object processing strategy prevails. ALOE's strengths mainly stem from its focused object-centric representations, fine-grained attention, along with self-supervised dynamics learning, which are particularly effective for the synthetic, structured object-oriented scenes where understanding the causal relationships and interactions between objects is crucial. In contrast, taking raw videos as input, MCG's sparse sampling approach with spatial-temporal blocks may overlook essential details pivotal for the comprehensive question-answering tasks that CLEVRER presents. Moreover, the cross-modal general knowledge that MCG acquired through large-scale pretraining on noisy video-text pairs might not transfer as effectively to the physical object-specific scenarios. 
These insights encourage future works toward evolving a more generalized model with interpretability and logical causal robustness behind question-answering ratiocination.

\subsection{Qualitative Results}
Figure \ref{fig:vis_exam} illustrates typical examples to qualitatively investigate the performance of the proposed MCG on both open-ended and multi-choice tasks. The top four present open-ended instances from ActivityNet-QA, including spatial relationship, temporal relationship, motion, and free tasks. To provide a comprehensive analysis of MCG's effectiveness and drawbacks, we combine the competitive SOTA model ALPRO (leveraging the same magnitude of external training data with MCG) for joint analysis. According to the results, we draw several important observations: 
1) Both ALPRO and MCG demonstrate the ability to comprehend questions of varying complexity and locate even small visual details, such as the watch in the first case and the nail in the second case.
2) MCG exhibits an advantage in generating longer answers, while ALPRO remains to face challenges in generating multi-word answers, as evident in the second and fourth cases.
3) The failure case highlights that MCG may still struggle to handle why-type questions or ambiguous answers. These observations further demonstrate that the difficulty of open-ended QA stems from question-answering reasoning and answer-generation challenges.

Furthermore, we present typical multi-choice examples from NExT-QA in the bottom half of Figure \ref{fig:vis_exam} to illustrate MCG's effectiveness and causal reasoning capabilities. Our observations from these visualizations are as follows:
1) MCG exhibits its potential to deliver reliable answers, even in the face of complex causal questions (fifth instance) or questions requiring common-sense comprehension (sixth instance).
2) For the tough description question with substantial background interference (seventh instance), our model demonstrates robustness by correctly distinguishing a tiny blue object from a sizeable yellow background.
3) As for the failure case, we find that both ALPRO and the proposed MCG inevitably suffer the unobservable and elusive confounding effect misled by the data bias, which leads the model to focus on the spurious correlations, e.g., connecting ``two ladies'' with ``pose for photographs'' in the last case. This phenomenon indicates the necessity of causality study in video question answering.
\section{Conclusion and Future Work}
We introduce MCG, an end-to-end multi-granularity contrastive cross-modal collaborative generative model for long-term VideoQA. MCG employs joint unimodal modeling to derive discriminative representations possessing high visual concepts and leverages a novel multi-granularity contrastive learning strategy to harness the intrinsically explicitly exhibited semantic correspondences. At its core, MCG formulates VideoQA as a generative task to reconcile existing discrepancies in VideoQA task formulations with a cross-modal collaborative generation module. It empowers MCG with the capability for cross-modal high-semantic fusion and generation to rationalize and answer, ensuring a more intuitive and effective approach to generating answers. MCG sets new benchmarks on four VideoQA datasets and shows strong generalization across diverse tasks.\\
\textbf{Limitations and Future Directions}
While MCG demonstrates robust performance in analyzing natural scenes, it encounters difficulties in synthetic scenes characterized by simple visuals yet complex dynamics, such as those found in the CLEVRER dataset. This challenge points to a broader need for enhanced video understanding and causal relationship reasoning. Moving forward, our future works will focus on developing a more generalized model with enhanced interpretability and logical causal robustness behind question-answering ratiocination. Moreover, we plan to leverage parameter-efficient tuning techniques to maximize VideoQA performance without proportionally increasing computational demands and explore transfer capabilities from large-scale foundation models to dynamic video scenarios, thereby enhancing the model’s ability to interpret complex temporal relationships.
\ifCLASSOPTIONcaptionsoff
  \newpage
\fi

\bibliographystyle{IEEEtran}
\bibliography{TIP2023}
\vspace{-6pt}
\end{document}